\documentclass[10pt,journal,letterpaper,compsoc,final]{IEEEtran} 
\usepackage{ifpdf}
\ifCLASSOPTIONcompsoc
\usepackage[nocompress]{cite}
\else
\usepackage{cite}
\fi

\usepackage[cmex10]{amsmath}
\usepackage{stfloats}

\usepackage{graphicx} 
\usepackage{subfigure} 

\usepackage[notheorems]{rr-math}
\usepackage{amsmath}

\usepackage{balance}

\usepackage[
textfont=sf,
labelfont={bf,sf},font=footnotesize,singlelinecheck=on]{caption} 

\usepackage{subfigure}
\usepackage{wrapfig}
\usepackage{lipsum,booktabs}

\newcommand{\data}[2]{{\mathsf{#1}\text{--}{\mathbf{\tt #2}}}} 

\newcommand{\datasmone}[1]{\ensuremath{\fontsize{6}{7.5}\selectfont \mathsf{#1}}} 
\newcommand{\dataName}[1]{\ensuremath{\fontsize{6}{7.5}\selectfont \mathsf{#1}}} 

\usepackage{booktabs}
\usepackage{comment}

\usepackage{multirow}
\usepackage{rotating}
\usepackage{subfigure}
\usepackage{epstopdf}
\usepackage{paralist}
\usepackage{url}
\usepackage{graphicx}
\usepackage{wrapfig}
\usepackage{xspace}
\usepackage[table]{xcolor}

\newcommand{\incomplete}[1]{}
\usepackage{tabularx}
\usepackage{nicefrac}
\usepackage{comment}

\usepackage{listings}
\usepackage{array}
\usepackage{ifpdf}
\usepackage{paralist}
\usepackage{multirow}
\usepackage{epsfig}
\usepackage{setspace}

\graphicspath{{./}{./images/}}
\newcolumntype{H}{>{\setbox0=\hbox\bgroup}c<{\egroup}@{}}
\definecolor{orange}{rgb}{1,0.5,0}
\definecolor{gray}{RGB}{20,20,20}
\definecolor{greencm}{RGB}{0,153,0}
\definecolor{verylightblue}{RGB}{191, 222, 253}

\newcommand{\verylightgreen}{green!10} 
\renewcommand{\verylightgreen}{blue!6}

\usepackage{algorithm}
\usepackage{algorithmicx}
\usepackage{algpseudocode}

\algblockdefx[parallel]{ParFor}{EndPar}[1][]{$\textbf{parallel for}$ #1 $\textbf{do}$}{$\textbf{end parallel}$}
\algrenewcommand{\alglinenumber}[1]{\fontsize{6.5}{7}\selectfont#1}
\algtext*{EndPar}

\algblockdefx[parallel]{parfor}{endpar}[1][]{$\textbf{parallel for}$ #1 $\textbf{do}$}{$\textbf{end parallel}$}
\algrenewcommand{\alglinenumber}[1]{\scriptsize#1:}

\algblockdefx[foreach1]{foreach}{endforeach}[1][]{$\textbf{for}$ #1 $\textbf{do}$}{$\textbf{end for}$}
\algrenewcommand{\alglinenumber}[1]{\scriptsize#1:}

\newcommand{\ra}[1]{\renewcommand{\arraystretch}{#1}} 

\algnotext{EndFor}
\algnotext{EndIf}
\algnotext{EndProcedure}

\ifpdf
\usepackage{epstopdf}
\fi
\graphicspath{{./}{./figures/}}

\graphicspath{{graphics/}{../graphics/}{./}}

\newcolumntype{P}[1]{>{\centering\arraybackslash}p{#1}}
\newcolumntype{M}[1]{>{\centering\arraybackslash}m{#1}}

\newcommand{\eol}{\end{enumerate}\setlength{\itemsep}{-\parsep}}

\relax

\newlength{\commentWidth}
\newcommand{\bspacing}{\begin{spacing}{1.4}}
\newcommand{\espacing}{\end{spacing}}

\definecolor{plotblue}{RGB}	{30,144,255}
\definecolor{plotgreen}{RGB}	{50,205,50}
\definecolor{plotred}{RGB}	{220,20,60}

\definecolor{myyellow}{RGB}{255,255,204}
\definecolor{myred}{RGB}{255,204,204}
\definecolor{myblue}{RGB}{204, 255, 255}
\definecolor{mygreen}{RGB}{204, 255, 204}

\definecolor{gray}{RGB}{150,150,150}

\definecolor{thedarkblue}{RGB}{0,0,120} 
\usepackage{hyperref}
\definecolor{mydarkblue}{rgb}{0,0.08,0.66}
\definecolor{mydarkblue}{rgb}{0,0.08,0.7}
\hypersetup{ 
pdftitle={Deep Feature Learning for Graphs},
pdfauthor={Ryan A. Rossi, Rong Zhou, and Nesreen K. Ahmed},
pdfsubject={Machine Learning},
pdfkeywords={Graph representation learning, relational function learning, transfer learning, feature learning, attributed graphs, deep learning, machine learning, representation learning, relational learning},
pdfborder=0 0 0,
pdfpagemode=UseNone,
colorlinks=true,
linkcolor=mydarkblue,
citecolor=mydarkblue,
filecolor=mydarkblue,
urlcolor=mydarkblue,
pdfview=FitH
}

\newcommand*\hrulefillvar[1][0.4pt]{\leavevmode\leaders\hrule height#1\hfill\kern0pt}

\newcommand{\be}{\begin{equation}}
\newcommand{\ee}{\end{equation}}
\newcommand{\bea}{\begin{eqnarray}}
\newcommand{\eea}{\end{eqnarray}}
\newcommand{\bit}{\begin{itemize}}
\newcommand{\eit}{\end{itemize}}

\definecolor{lightgray}{rgb}{0.93,0.93,0.93}

\definecolor{lightblue}{rgb}{0.5,0.90,1.0}
\definecolor{lightgreen}{rgb}{0.5,0.92,0.5}
\definecolor{lightred}{rgb}{0.98,0.5,0.5}
\definecolor{lightyellow}{rgb}{1,0.90,0.40}

\newcommand\TTT{\rule{0pt}{3.2ex}}

\newcommand\BB{\rule[-1.0ex]{0pt}{0pt}}

\definecolor{myyellow}{RGB}{255,255,204}
\definecolor{myred}{RGB}{255,204,204}
\definecolor{myblue}{RGB}{0,200,255}
\definecolor{mygreen}{RGB}{80,220,80}

\newcommand{\eg}{\emph{e.g.}}
\newcommand{\ie}{\emph{i.e.}}
\newcommand{\iid}{\emph{i.i.d.}}

\newcommand{\wrt}{\emph{w.r.t.}}

\newcommand{\abs}[1]{\left|#1\right|}

\newcommand{\inner}[2]{\langle #1,#2 \rangle}
\newcommand{\innerS}[1]{\langle #1 \rangle}

\algblockdefx[parallel]{parallelfor}{parallelend}
[1][]{\textbf{parallel for} #1}
{\textbf{end parallel}}

\newcommand{\ds}\displaystyle
\newcommand{\mbb}\mathbb
\newcommand{\mc}\mathcal
\newcommand{\del}\nabla

\newcommand{\beqstar}{\begin{eqnarray*}}
\newcommand{\eeqstar}{\end{eqnarray*}}

\definecolor{thegreen}{rgb}{0,.5,0}
\definecolor{idea}{rgb}{0,.6,0.1}
\definecolor{problem}{rgb}{0.7,0,0.1}
\definecolor{comment-green}{rgb}{0,.3,0}
\definecolor{theblue}{rgb}{0,0,.8}
\definecolor{light-gray}{gray}{0.98}
\definecolor{comment-color}{rgb}{0,0,.8}
\definecolor{string-color}{rgb}{0,.75,0}
\definecolor{border-blue}{rgb}{0,0,.6}



\algblockdefx[fornew1]{foreach}{endforeach}
[1][]{\textbf{for} #1}
{\textbf{end for}}

\newcommand{\setAlgFontSize}{\fontsize{8pt}{9pt}\selectfont} 

\newcommand{\multilinenospace}[1]{\State \parbox[t]{\dimexpr\linewidth-\algorithmicindent}{\begin{spacing}{1.1}\setAlgFontSize#1\strut \end{spacing}}}

\newcommand{\deepGL}{\ensuremath{\mathsf{DeepGraF}}}
\renewcommand{\deepGL}{\ensuremath{\mathrm{DeepGL}}}
\newcommand{\deepGLbold}{\ensuremath{\mathrm{\bf DeepGL}}}
\newcommand{\deepGLsf}{\ensuremath{\mathsf{DeepGL}}}

\providecommand{\RR}{\mathbb{R}} 
 
\providecommand{\Std}{\mathbb{S}}

\renewcommand{\argmax}{\operatornamewithlimits{\arg \; \max}}

\renewcommand{\phi}{\ensuremath{\Phi}}

\providecommand{\mPhi}{\ensuremath{{\rm \boldsymbol\Phi}}}

\providecommand{\maxiter}{\ensuremath{\mathrm{T}}}
\providecommand{\ops}{\ensuremath{\mPhi}}

\providecommand{\F}{\ensuremath{\mathcal{F}}} 
\providecommand{\corr}{\ensuremath{\mathbb{K}}}

\providecommand{\E}{\ensuremath{E}}

\providecommand{\p}{\ensuremath{p}}

\providecommand{\p}{\ensuremath{p}}

\renewcommand{\bigO}[1]{\ensuremath{\mathcal{O}(#1)}}

\providecommand{\rsm}{\ensuremath{\textsc{rsm}}}

\providecommand{\data}[2]{{\mathsf{#1}\text{--}{\mathbf{\tt #2}}}} 

\providecommand{\datasmone}[1]{{\fontsize{6}{7.5}\selectfont \textbf{#1}}}

\providecommand{\m}{\ensuremath{M}} 
\providecommand{\n}{\ensuremath{N}}  
\providecommand{\nF}{\ensuremath{F}} 
  
\renewcommand{\d}{\ensuremath{d}}  
\providecommand{\f}{\ensuremath{f}}

\providecommand{\N}{\ensuremath{\Gamma}}

\usepackage{comment}

\usepackage{setspace}

\graphicspath{{./}{./images/}}
\newcolumntype{H}{>{\setbox0=\hbox\bgroup}c<{\egroup}@{}}

\usepackage{url}

\usepackage{amssymb}
\usepackage{graphicx}

\usepackage{url}

\begin{document} 

\title{Deep Feature Learning for Graphs}

\author{Ryan A. Rossi, Rong Zhou, and Nesreen K. Ahmed
\IEEEcompsocitemizethanks{
\IEEEcompsocthanksitem R. A. Rossi and R. Zhou are with Palo Alto Research Center (Xerox PARC), 
3333 Coyote Hill Rd, Palo Alto, CA USA\protect\\
E-mail: rrossi@parc.com, rzhou@parc.com
\IEEEcompsocthanksitem N. K. Ahmed is with Intel Labs, 3065 Bowers Ave, Santa Clara, CA USA\protect\\
E-mail: nesreen.k.ahmed@intel.com
}
\thanks{}}

\markboth{}{Rossi \MakeLowercase{\textit{et al.}}: Deep Feature Learning for Graphs}

\IEEEcompsoctitleabstractindextext{
\begin{abstract}
This paper presents a general graph representation learning framework called $\deepGLsf$ for learning deep node \emph{and} edge representations from large (attributed) graphs.
In particular, $\deepGLsf$ begins by deriving a set of base features (\eg, graphlet features) and automatically learns a multi-layered hierarchical graph representation where each successive layer leverages the output from the previous layer to learn features of a higher-order.
Contrary to previous work, $\deepGLsf$ learns \emph{relational functions} (each representing a feature) that generalize across-networks and therefore useful for graph-based transfer learning tasks.
Moreover, $\deepGLsf$ naturally supports attributed graphs, learns interpretable graph representations, and is space-efficient (by learning sparse feature vectors).
In addition, $\deepGLsf$ is expressive, flexible with many interchangeable components, efficient with a time complexity of $\mathcal{O}(|E|)$, and scalable for large networks via an efficient parallel implementation.
Compared with the state-of-the-art method, $\deepGLsf$ is 
(1) \textbf{effective} for across-network transfer learning tasks \emph{and} attributed graph representation learning,
(2) \textbf{space-efficient} requiring up to 6$\times$ less memory, 
(3) \textbf{fast} with up to 182$\times$ speedup in runtime performance, and 
(4) \textbf{accurate} with an average improvement of 20$\%$ or more on many learning tasks.
\end{abstract}

}

\maketitle
\IEEEdisplaynotcompsoctitleabstractindextext
\IEEEpeerreviewmaketitle

\section{Introduction} \label{sec:intro}
\PARstart{L}{earning} a useful graph representation 
lies at the heart \emph{and} success of many within-network \emph{and} across-network machine learning tasks such as
node and link classification~\cite{neville2000iterative,sen2008collective,mcdowell2009cautious,rossi2012dynamic-srl,vishwanathan2010graph},
anomaly detection~\cite{noble2003graph,akoglu2015graph,rossi2011modeling}, 
link prediction~\cite{al2011survey,getoor:icdmw07},
dynamic network analysis~\cite{nicosia2013graph,kovanen2011temporal},
community detection~\cite{radicchi2004defining,palla2005uncovering}, 
role discovery~\cite{rossi2015-tkde,borgatti1992notions,airoldi2008mixed},
visualization and sensemaking~\cite{gvis-icwsm15,pienta2015scalable,fang2017carina}, 
network alignment~\cite{koyuturk2006pairwise}, and many others.
Indeed, the success 
of machine learning methods largely depends on data representation~\cite{goodfellow2016deep,rossi12jair}.
Methods capable of learning such representations have many advantages over feature engineering in terms of cost and effort.
The success of graph-based machine learning algorithms depends largely on data representation.
For a survey and taxonomy of relational representation learning, see~\cite{rossi12jair}.

Recent work has largely been based on the popular skip-gram model~\cite{skipgram} originally introduced for learning vector representations of words in the natural language processing (NLP) domain.
In particular, DeepWalk~\cite{deepwalk} applied the successful word embedding framework from~\cite{word2vec} (called word2vec) to embed the nodes such that the co-occurrence frequencies of pairs in short random walks are preserved. 
More recently, node2vec~\cite{node2vec} introduced hyperparameters to DeepWalk that tune the depth and breadth of the random walks.
These approaches have been extremely successful and have shown to outperform a number of existing methods on tasks such as node classification.

However, much of this past work has focused on \emph{node features}~\cite{deepwalk,node2vec,line}.
These node features provide only a coarse representation of the graph.
Existing methods are also unable to leverage attributes (\eg, gender, age) and lack support for typed graphs.
In addition, features from these methods do \emph{not} generalize to other networks and thus are unable to be used for across-network transfer learning tasks.
Existing methods are also not space-efficient as the node feature vectors are completely dense.
For large graphs, the space required to store these dense features can easily become too large to fit in-memory.
The features are also notoriously difficult to interpret and explain which is becoming increasingly important in practice~\cite{vellido2012making}.
Furthermore, existing embedding methods are also
unable to capture higher-order subgraph structures as well as learn a hierarchical graph representation from such higher-order structures.
Finally, these methods are also inefficient with runtimes that are orders of magnitude slower than the algorithms presented in this paper
(as shown later in Section~\ref{sec:exp}).

\begin{figure*}[t!]
\centering
\includegraphics[width=0.65\linewidth]{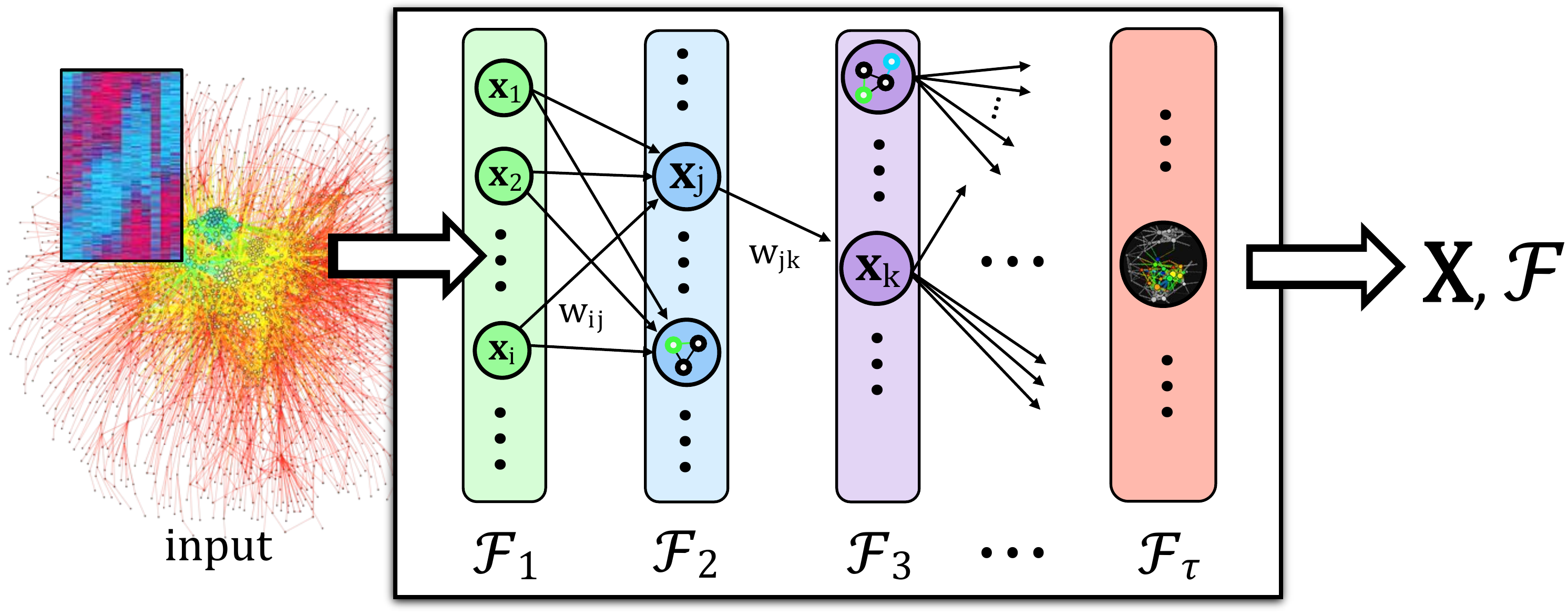}
\vspace{1mm}
\caption{Overview of the $\deepGLsf$ architecture for graph representation learning. 
Let $\mW= \big[w_{ij}\big]$ be a matrix of feature weights where $w_{ij}$ (or $W_{ij}$) is the weight 
between the feature vectors $\vx_i$ and $\vx_j$.
Notice that $\mW$ has the constraint that $i<j<k$ and $\vx_i$, $\vx_j$, and $\vx_k$ are increasingly deeper.
It is straightforward to see that $\mathcal{F} = \F_{1} \cup \F_{2} \cup \cdots \cup \F_{\tau}$, and thus, $\abs{\mathcal{F}} = |\F_{1}| + |\F_{2}| + \cdots + |\F_{\tau}|$.
Moreover, the layers are ordered where $\F_{1} < \F_{2} < \cdots < \F_{\tau}$ 
such that if $i<j$ then $\F_{j}$ is said to be a deeper layer $\wrt$ $\F_{i}$.
See Table~\ref{table:notation} for a summary of notation.
}
\label{fig:framework}
\end{figure*}

In this work, we present a general, expressive, and flexible \emph{deep graph representation learning framework} called $\deepGL$ that overcomes many of the above limitations.
Intuitively, $\deepGL$ begins by deriving a set of base features using the graph structure, attributes, and/or both.
The base features are iteratively composed using a set of learned \emph{relational feature operators} (Fig.~\ref{table:rel-deep-learning-operators}) 
that operate over the feature values of the (distance-$\ell$) neighbors of a graph element (node, edge; see Table~\ref{table:notation}) to derive higher-order features from lower-order ones
forming a hierarchical graph representation where each layer consists of features of increasingly higher orders.
At each feature layer, 
$\deepGL$ searches over a space of relational functions
defined compositionally in terms of a set of \emph{relational feature operators} applied to each feature given as output in the previous layer.
Features (or relational functions) are retained if they are novel and thus add important information that is not captured by any other feature in the set.
See below for a summary of the advantages and properties of $\deepGL$.

\subsection{Summary of Contributions} \label{sec:intro-summary}
\noindent
The proposed framework, $\deepGL$, overcomes many limitations of existing work and has the following key properties:
\begin{itemize}
\item \textbf{Novel framework}: This paper presents a deep hierarchical graph representation learning framework called $\deepGL$ for
large (attributed) networks that generalizes for discovering both node and edge features.
The framework is flexible with many interchangeable components, expressive, and shown to be effective for a wide variety of applications.

\item \textbf{Attributed graphs}: 
$\deepGL$ is naturally able to learn graph representations from both attributes (if available) \emph{and} the graph structure.

\item \textbf{Graph-based transfer learning}:
Contrary to existing work, $\deepGL$ 
naturally supports 
across-network transfer learning tasks
as it learns relational functions that generalize for computation on any arbitrary graph.

\item \textbf{Sparse feature learning}: 
It is space-efficient by learning a sparse graph representation that requires up to $6$x less space than existing work.

\item \textbf{Interpretable and Flexible}:
Unlike embedding methods, 
$\deepGL$ learns interpretable and explainable features. $\deepGL$ is also flexible with many interchangeable components making it well-suited for a variety of applications, graphs, and learning scenarios.

\item \textbf{Hierarchical graph representation}:
$\deepGL$ learns hierarchical graph representations 
where each successive layer uses the output from the previous layer 
to derive features of a higher-order.

\item \textbf{Higher-order structures}:
Features based on higher-order structures are learned from lower-order subgraph features via propagation.
This is in contrast to existing methods that
are unable to capture such higher-order subgraph structures.

\item \textbf{Efficient, Parallel, and Scalable}:
It is fast with a runtime that is linear in the number of edges. It scales to large graphs via a simple and efficient parallelization.
Notably, strong scaling results are observed in Section~\ref{sec:exp}.
\end{itemize}

\section{Related Work} \label{sec:related-work}
In this section, we highlight how $\deepGL$ differs from related work.

\smallskip
\noindent
\textbf{Node embedding methods}:
There has been a lot of interest recently in learning a set of useful features from large-scale networks automatically~\cite{node2vec,deepwalk,line}. In particular, recent methods that apply the popular word2vec framework to learn node embedding~\cite{node2vec,deepwalk}.
The proposed $\deepGL$ framework differs from these methods 
in six fundamental ways:
(1) It naturally supports attributed graphs
(2) Learns complex relational functions that transfer for across-network learning.
(3) $\deepGL$ learns important and useful edge \emph{and} node representations,
whereas existing work is limited to \emph{node features}~\cite{deepwalk,node2vec,line}.
(4) It learns sparse features and thus extremely space-efficient for large networks.
(5) It is fast and efficient with a runtime that is linear in the number of edges.
(6) It is also completely parallel and shown in Section~\ref{sec:exp} to scale strongly. 
Other key differences are summarized previously in Section~\ref{sec:intro}.

\smallskip
\noindent\textbf{Higher-order network analysis}: 
Other methods use high-order network properties (such as graphlet frequnecies) as features for graph classification~\cite{vishwanathan2010graph}. Graphlets are small induced subgraphs and have been used for graph classification~\cite{vishwanathan2010graph}
and visualization and exploratory analysis~\cite{pgd}. 
However, our work focuses on using graphlets counts as base features for learning node and edge representations from large networks.
Furthermore, previous feature learning methods are typically based on random walks or limited to features based on simple degree and egonet-based features. 
Thus, another contribution and key difference between existing approaches is the use of higher-order network motifs (based on small k-vertex subgraph patterns called graphlets) for feature learning and extraction.
To the best of our knowledge, this paper is the first to use network motifs (including all motifs of size 3, 4, and 5 vertices) as base features for graph representation learning.

\smallskip
\noindent\textbf{Sparse graph feature learning}: 
This work proposes the first practical space-efficient approach that learns sparse node/edge feature vectors.
Notably, $\deepGL$ requires significantly less space than existing node embedding methods~\cite{deepwalk,node2vec,line} (see Section~\ref{sec:exp}).
In contrast, existing embedding methods store completely dense feature vectors 
which is impractical for any relatively large network, \eg, 
they require more than 3TB of memory for a 750 million node graph with 1K features.

\begin{table}
\caption{Summary of notation}
\centering 
\small
\fontsize{8}{8.5}\selectfont
\setlength{\tabcolsep}{6pt} 
\label{table:notation}
\def\arraystretch{1.28}
\begin{tabularx}{1.0\linewidth}
{@{}rX@{}} 
\toprule
$G$ & (un)directed (attributed) graph\\
$\mA$ & sparse adjacency matrix of the graph $G=(V,E)$ \\
$\n, \m$ & number of nodes and edges in the graph \\
$\nF, L$ & number of learned features and layers  \\
$\mathcal{G}$ & set of graph elements $\{g_1,g_2,\cdots\}$  (nodes, edges)\\
$\d^{+}_v$, $\d^{-}_{v}$, $\d_{v}$ & outdegree, indegree, degree of vertex $v$ \\
$\N^{+}_{}\!(g_i)$, $\N^{-}_{}\!(g_i)$ & out/in neighbors of graph element $g_i$\\
$\N_{}(g_i)$ & neighbors (adjacent graph elements) of $g_i$\\
$\N_{\ell}(g_i)$ & $\ell$-neighborhood $\N(g_i) = \{g_j \in \mathcal{G} \,|\, \mathrm{dist}(g_i, g_j) \leq \ell \}$ \\ 
$\mathrm{dist}(g_i, g_j)$ &  shortest distance between $g_i$ and $g_j$ \\
$S$ & set of graph elements related to $g_i$, \eg, $S=\N_{}(g_i)$ \\
$\mX$  & a feature matrix \\ 
$\vx$ & an $\n$ or $\m$-dimensional feature vector \\
$\mX_{\tau}$ & (sub)matrix of features from layer $\tau$ \\
$\bar{\mX}$ & diffused feature vectors $\bar{\mX} = [ \bar{\vx}_1 \;\; \bar{\vx}_2 \;\; \cdots ]$ \\
$\abs{\mX}$ & number of nonzeros in a matrix $\mX$ \\
$\F$ & set of \emph{feature definitions/functions} from $\deepGL$\\
$\F_k$ & $k$-th feature layer (where $k$ is the depth) \\
$\f_i$ & relational function (definition) of $\vx_i$ \\
$\mPhi$ & relational operators $\mPhi = \{\phi_{1}, \cdots, \phi_{K}\}$\\
$\mathbb{K}(\cdot)$ & a feature evaluation criterion \\
$\lambda$ & tolerance/feature similarity threshold \\
$\alpha$ & transformation hyperparameter \\
$\vx^{\prime} = \phi_i\innerS{\vx}$ & relational operator applied to each graph element \\
\bottomrule
\end{tabularx}
\end{table}

\section{Framework} \label{sec:framework}
This section presents the $\deepGL$ framework.
Since the framework naturally generalizes for learning node and edge representations, 
it is described generally for a set of graph elements (\eg, nodes or edges).\footnote{For convenience, $\deepGL$-edge and $\deepGL$-node are sometimes used to refer to the edge and node representation learning variants of $\deepGL$, respectively.}
An overview of the $\deepGL$ architecture is provided in Fig.~\ref{fig:framework}.
A summary of notation is provided in Table~\ref{table:notation}.

\subsection{Base Graph Features} \label{sec:framework-base-features}
The first step of $\deepGL$ (Alg.~\ref{alg:framework}) is to derive a set of \emph{base graph features}\footnote{The term \emph{graph feature} refers to an edge or node feature; and includes features derived by meshing the graph structure with attributes.} using the graph topology and attributes (if available).
Note that $\deepGL$ generalizes for use with an arbitrary set of base features, and thus it is not limited to the base features discussed below.
Given a graph $G=(V,E)$, we first decompose $G$ into its smaller subgraph components called graphlets (network motifs)~\cite{pgd} using local graphlet decomposition methods~\cite{rossi17graphlet-est,ahmed16bigdata} 
and append these features to $\mX$.
This work derives such features by counting all node or edge \emph{orbits} with up to $4$ and/or $5$-vertex graphlets.
Orbits (graphlet automorphisms) are counted for each node or edge in the graph based on whether a node or edge representation is warranted (as our approach naturally generalizes to both).
Note there are 15 node and 12 edge orbits with 2-4 nodes;
and 73 node and 68 edge orbits with 2-5 nodes.
However, $\deepGL$ trivially handles other types of subgraph (graphlet) sizes and features including graphlets that are directed/undirected, typed/heterogeneous, and/or temporal.
Furthermore, one can also derive such subgraph features efficiently by leveraging fast and accurate graphlet estimation methods (\eg,~\cite{rossi17graphlet-est,ahmed16bigdata}).

\begin{figure*}[t!]
\centering
\begin{minipage}{0.42\textwidth}
\scalebox{0.9}{
\setlength{\tabcolsep}{6pt} 
\centering 
\small
\def\arraystretch{1.5}
\begin{tabularx}{1.0\linewidth}{@{}rH X HH@{}} 
\textbf{Operator} & 
\textbf{Symbol} &
\textbf{Definition} & \textbf{Definition} \\
\midrule
\def\arraystretch{1.6}
Hadamard 	& $\boxplus$ & 
$\phi\innerS{S, \vx} = \prod\limits_{s_j \in S} x_{j}$ \\ 

mean		& $\boxdot$  & 
$\phi\innerS{S, \vx} = \frac{1}{|S|} \sum\limits_{s_j \in S} x_{j}$ &
\\

sum	    & $\otimes$ & 
$\phi\innerS{S, \vx} = \sum\limits_{s_j \in S} x_{j}$ &
\\

maximum    &  $\max$	&
$\phi\innerS{S, \vx} = \max\limits_{s_j \in S} \; x_{j}$ &
\\

Weight. $L^p$    &   $\|\cdot\|_{\bar{p}}$ & 
$\phi\innerS{S, \vx} = \sum\limits_{s_j \in S} \abs{x_{i} - x_{j}}^{p}$ & 
\\

RBF & $-$ &
$\phi\innerS{S, \vx} =$ 
\fontsize{8}{7.5}\selectfont
$\exp\Big( - \frac{1}{\sigma^2} \sum\limits_{s_j \in S} \big[x_{i} - x_{j}\big]^{2}\Big)$ &
\\

\midrule
\end{tabularx}
}
\end{minipage}
\begin{minipage}[b!]{0.24\textwidth}
\includegraphics[width=0.7\linewidth]{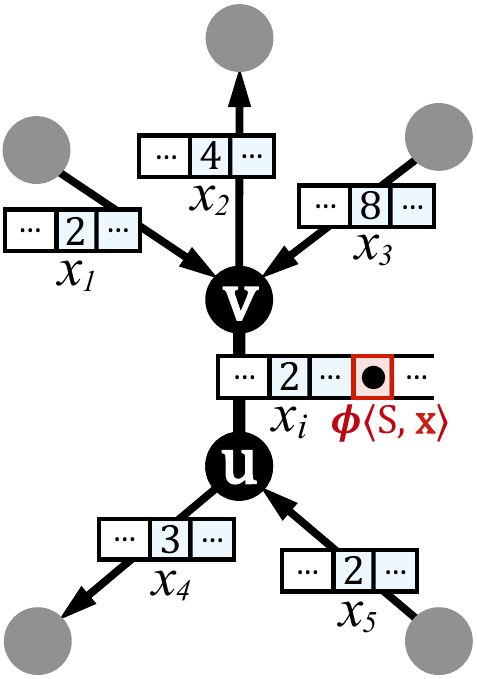}
\end{minipage}
\caption{
Relational feature operators.
Left:
Summary of a few \emph{relational feature operators}.
Note that $\deepGLsf$ is flexible and generalizes to any arbitrary set of relational operators.
The set of relational feature operators can be learned via a validation set.
Recall the notation from Table~\ref{table:notation}.
For generality, $S$ is defined in Table~\ref{table:notation} as a set of related graph elements (nodes, edges) of $g_i$ and thus $s_j \in S$ may be an edge $s_j = e_j$ or a node $s_j = v_j$; in this work $S \in \big\lbrace\N_{\ell}(g_i),\, \N^{+}_{\ell}\!(g_i),\, \N^{-}_{\ell}\!(g_i)\big\rbrace$ (Alg.~\ref{alg:feature-layer}).
The relational operators generalize easily for $\ell$-distance neighborhoods (\eg, $\N_{\ell}(g_i)$
where $\ell$ is the distance).
Right:
An intuitive example for an edge $e=(v,u)$ and a relational operator $\phi\in\mPhi$.
Suppose $\phi = $ relational sum operator and $S=\{e_1, e_2, e_3, e_4, e_5\}=\N_{\ell}(e_i)$ where $\ell=1$ (distance-1 neighborhood), then $\phi\innerS{S, \vx}=19$.
Now, suppose $S=\{e_2, e_4\}=\N_{\ell}^{+}\!(e_i)$ then $\phi_p\innerS{S, \vx}=7$ \emph{and} similarly, if $S=\{e_1, e_3, e_5\}=\N_{\ell}^{-}\!(e_i)$ then $\phi\innerS{S, \vx}=12$. 
Note $\vx = \big[ \;\; x_1 \;\; x_2 \;\; \cdots \;\; x_i \;\; \cdots \;\; \big] \in \RR^{\m}$ where $x_i$ is the $i$-th element of $\vx$ for edge $e_i$.
Notice that $\phi\innerS{S, \vx}$ refers to the application of $\phi$ to $S$ for a single edge $e=(v,u)$. 
For simplicity, we also use $\phi\innerS{\vx}$ (whenever clear from context) to refer to the application of $\phi$ to all sets $S$ derived from each graph element in $G$ (and thus the output of $\phi\innerS{\vx}$ in this case is a feature vector with a single feature-value for each graph element).
As an example, suppose $S = \N_{\ell}(e_i)$ where $\ell=1$ (distance-1 neighborhood), then one can view $\vx^{\prime} = \phi\innerS{\vx}$ 
as $\big[ \, \phi\innerS{\N_{\ell}(e_1), \vx} \; \cdots \; \phi\innerS{\N_{\ell}(e_{\m}), \vx} \, \big]$ where $S$ for each $e_i \in E$ has been replaced with the set of in/out neighbors for each $e_i \in E$ denoted $\N_{\ell}(e_i)$. 
}
\label{table:rel-deep-learning-operators}
\end{figure*}

\begin{figure}[th!]
\centering
\subfigure[External egonet features]{\includegraphics[width=0.40\linewidth]{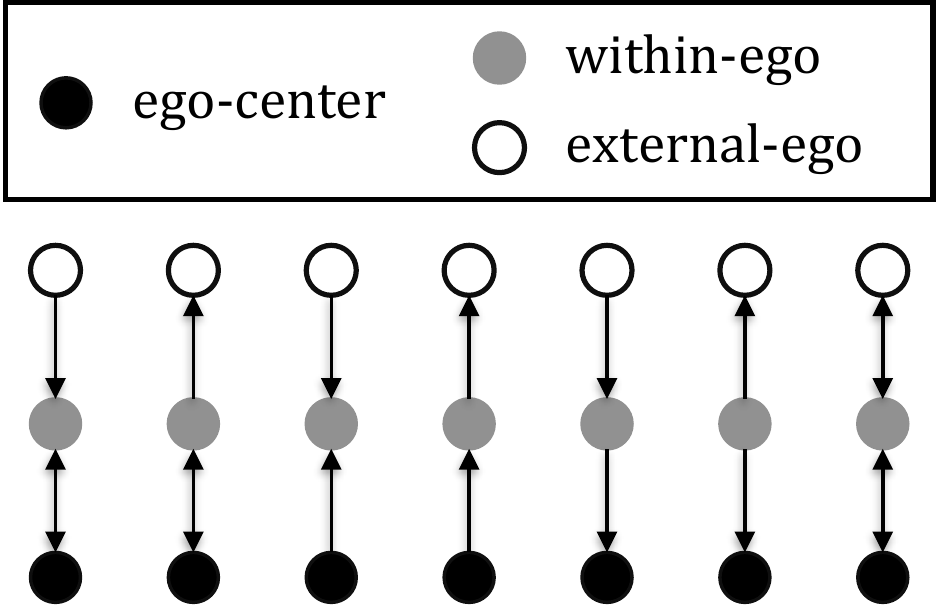}\label{fig:external-egonet-pattens}}
\hfill
\subfigure[Within egonet features]{\includegraphics[width=0.55\linewidth]{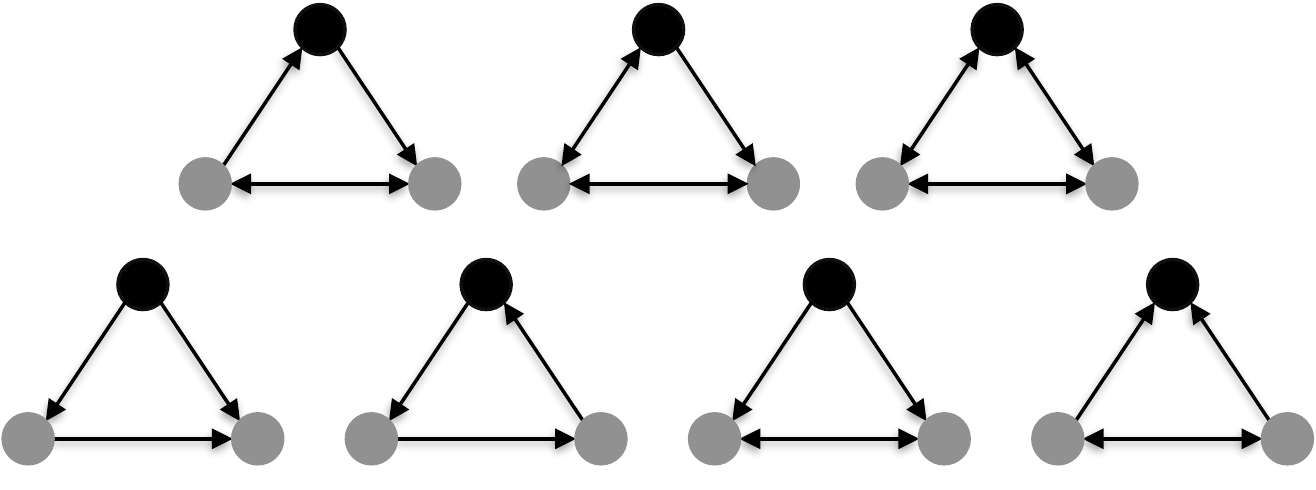}\label{fig:within-egonet-pattens}}
\caption{\textbf{Egonet Features}. 
The set of base ($\ell$=1 hop)-egonet graph features.
\text{(a)} the external egonet features; 
\text{(b)} the within egonet features.
Note that it is straightforward to generalize these egonet features to edges.
The $\deepGL$ framework naturally supports other base features as well.
See the legend for the vertex types: ego-center ({\large  $\bullet$}), within-egonet vertex (\textcolor{gray}{\large $\bullet$}), and external egonet vertices (\textcolor{black}{\large $\circ$}).
}
\label{fig:egonet-external-and-within-features}
\end{figure}

We also derive simple base features such as in/out/total/weighted degree and k-core numbers for each graph element (node, edge) in $G$.
For edge feature learning we derive edge degree features for each edge $(v,u) \in E$ and each $\circ  \in \{+, \times\}$ as follows:
\[
\big[
\begin{matrix}
\,
\d^{+}_v \circ \d^{+}_{u}, \,\,\, &
\d^{-}_v \circ \d^{-}_{u}, \,\,\, &
\d^{-}_v \circ \d^{+}_{u}, \,\,\, &
\d^{+}_v \circ \d^{-}_{u}, \,\,\, &
\d_v \circ \d_u\,\,
\\
\end{matrix}\big]
\]
where $\d_v = \d^{+}_v \circ \d^{-}_v$ 
and recall from Table~\ref{table:notation} that $\d^{+}_v$, $\d^{-}_{v}$, and $\d_v$ denote the out/in/total degree of $v$.
In addition, egonet features are also used.
The external and within-egonet features for nodes are provided in Fig.~\ref{fig:egonet-external-and-within-features} and used as base features in $\deepGL$-node.
It is straightforward to extend these egonet features to edges for learning edge representations.
For all the above base features, we also derive variations based on direction (in/out/both) and weights (weighted/unweighted).
Observe that $\deepGL$ naturally supports many other graph properties including efficient/linear-time properties such as PageRank.
Moreover, fast approximation methods with provable bounds can also be used to derive features such as the 
local coloring number 
and largest clique 
centered at the neighborhood of each graph element (node, edge) in $G$.

A key advantage of $\deepGL$ lies in its ability to naturally handle attributed graphs. 
We discuss the four general cases below that include learning a node or edge feature-based representation given an initial set of node or edge attributes.
For learning a node representation (via $\deepGL$-node) given $G$ and an initial set of \emph{edge attributes}, we simply derive node features by applying the set of relational feature operators (Fig.~\ref{table:rel-deep-learning-operators}) to each edge attribute.
Conversely, learning an edge representation ($\deepGL$-edge) given $G$ and an initial set of \emph{node attributes}, we derive edge features by applying each relational operator $\phi \in \mPhi$ to the nodes at either end of the edge\footnote{Alternatively, each relational operator $\phi \in \mPhi$ can be applied to the various combinations of in/out/total neighbors of each pair of nodes i and j that form an edge.}.
Finally, when the input attributes match the type of graph element (node, edge) for which a feature representation is learned, then the attributes are simply appended to the feature matrix $\mX$.

\subsection{Space of Relational Functions and Expressivity}
\label{sec:framework-space-of-relational-functions}
In this section, we formulate the space of relational functions\footnote{The terms graph function and relational function are used interchangeably} that can be expressed and searched over by $\deepGL$. 
Recall that unlike recent node embedding methods~\cite{deepwalk,node2vec,line}, the proposed approach learns graph functions that are transferable across-networks for a variety of important graph-based transfer learning tasks such as 
\emph{across-network} 
prediction,
anomaly detection,
graph similarity, 
matching, 
among others.

\subsubsection{Composing Relational Functions} \label{sec:composing-relational-functions}
The space of relational functions searched via $\deepGL$ is defined \emph{compositionally} in terms of a set of \emph{relational feature operators} $\mPhi=\{\phi_1,\cdots,\phi_K\}$.\footnote{Note $\deepGL$ may also leverage traditional feature operators used for $\iid$ data.}
A few relational feature operators are provided in Fig.~\ref{table:rel-deep-learning-operators}; see~\cite{rossi12jair} (pp. 404) for a wide variety of other useful relational feature operators.
The expressivity of $\deepGL$ (\ie, space of relational functions expressed by $\deepGL$) depends on a few flexible and interchangeable components including: 
(\emph{i}) the initial base features (derived using the graph structure, initial attributes given as input, or both),
(\emph{ii}) a set of \emph{relational feature operators} $\mPhi=\{\phi_1,\cdots,\phi_K\}$, 
(\emph{iii}) the sets of ``related graph elements'' $S \in \mathcal{S}$ (\eg, the in/out/all neighbors within $\ell$ hops of a given node/edge) that are used with each relational feature operator $\phi_p \in \mPhi$, and finally, 
(\emph{iv}) the number of times each relational function is composed with another (\ie, the depth). 
Intuitively, observe that under this formulation each feature vector $\vx^{\prime}$ from $\mX$ (that is not a base feature) can be written as a composition of relational feature operators applied over a base feature.
For instance, given an initial \emph{base feature} $\vx$, let $\vx^{\prime} = \phi_k(\phi_j(\phi_i\innerS{\vx})) = (\phi_k \circ\, \phi_j \circ\, \phi_i)(\vx)$ 
be a feature vector given as output by applying the relational function constructed by composing the \emph{relational feature operators} $\, \phi_k \circ\, \phi_j \circ\, \phi_i$.
Obviously, more complex relational functions are easily expressed such as those involving compositions of different relational feature operators (and possibly different sets of related graph elements).
Furthermore, as illustrated in Fig.~\ref{fig:framework}, $\deepGL$ is able to learn relational functions that often correspond to increasingly higher-order subgraph features based on a set of initial lower-order (base) subgraph features (typically all 3, 4, and/or 5 vertex subgraphs).
Intuitively, just as filters are used in Convolutional Neural Networks (CNNs)~
\cite{goodfellow2016deep}, one can think of $\deepGL$ in a similar way, but instead of simple filters, we have features derived from lower-order subgraphs being combined 
in various ways to capture higher-order subgraph patterns of increasingly complexity at each successive layer.

\subsubsection{Summation and Multiplication}
\label{sec:summation-and-multiplication}
We can also derive a wide variety of functions compositionally by adding and multiplying relational functions (\eg, $\phi_i + \phi_j$, and $\phi_i \times \phi_j$).
A \emph{sum of relational functions} is similar to an OR operation in that two instances are ``close'' if either has a large value, and similarly, a \emph{product of relational functions} is analogous an AND operation as two instances are close if both relational functions have large values.

\begin{figure*}[t!]
\centering
\begin{minipage}{0.82\textwidth}
\algblockdefx[parallel]{parfor}{endpar}[1][]{$\textbf{parallel for}$ #1 $\textbf{do}$}{$\textbf{end parallel}$}
\algtext*{endpar}
\algrenewcommand{\alglinenumber}[1]{\fontsize{6.5}{9}\selectfont#1:}
\begin{algorithm}[H]
\caption{\,\small \fontsize{9}{10.5}\selectfont
The $\deepGLbold$ {\bf framework} for learning deep graph representations (node/edge features) 
from (attributed) graphs where the features are expressed as relational functions that naturally transfer across-networks.
\vspace{0.5mm}
}
\label{alg:framework}
{
\begin{spacing}{1.15}
\fontsize{7}{8}\selectfont
\begin{algorithmic}[1]
\fontsize{8}{9}\selectfont
\vspace{0.5mm}
\Require ~\newline
\quad\quad 
a directed and possibly weighted/labeled/attributed graph $G = (V,E)$ 
\newline\quad\quad
a set of relational feature operators $\mPhi=\{\phi_1, \cdots, \phi_K\}$ (Fig.~\ref{table:rel-deep-learning-operators})
\newline\quad\quad
a feature evaluation criterion $\corr\inner{\cdot}{\cdot}$ 
\newline\quad\quad
an upper bound on the number of feature layers to learn $\maxiter$
\medskip
\fontsize{8}{9}\selectfont

\multilinenospace{
Given $G$ and $\mX$, construct base features (see text for further details) 
and add the feature vectors to $\mX$ and definitions to $\F_{1}$; and set $\F \leftarrow \F_{1}$.
\label{algline:base-features}}
\vspace{1.0mm}

\State Transform \emph{base feature vectors} (if warranted); Set $\tau \leftarrow 2$ \label{algline:transform-base-features-logarthmic-binning}
\vspace{1.0mm}

\Repeat \label{algline:repeat-feature-layer} \Comment{feature layers $\F_{\tau}$ for $\tau = 2,...,\maxiter$}
\vspace{0.6mm}
\multilinenospace{
Search the space of features defined by applying relational feature operators $\mPhi=\{\phi_1,\cdots,\phi_K\}$ to features $\big[\; \cdots \;\; \vx_{i} \;\; \vx_{i+1} \;\; \cdots \;\big]$ given as output in the previous layer $\F_{\tau-1}$ (via Alg.~\ref{alg:feature-layer}).
Add feature vectors to $\mX$ and functions/def. to $\F_{\tau}$.
\label{algline:construct-candidates-and-search-features}
}
\vspace{0.01mm}

\State Transform feature vectors of layer $\F_{\tau}$ (if warranted) \label{algline:transform-feature-layer-logarthmic-binning}
\vspace{0.8mm}

\multilinenospace{
Evaluate the features (functions) in layer $\F_{\tau}$ using the criterion $\mathbb{K}$ to score feature pairs along with a feature selection method to select a subset (\eg, see Alg.~\ref{alg:score-and-prune-feature-layer}).
\label{algline:main-prune-feature-layer}
}

\vspace{0.8mm}
\multilinenospace{Discard features from $\mX$ that were pruned (not in $\F_{\tau}$) and set $\F \leftarrow \F \cup \F_{\tau}$
\label{algline:removed-pruned-feats-from-X}
}

\vspace{0.7mm}
\multilinenospace{Set $\tau \leftarrow \tau + 1$ and initialize $\F_{\tau}$ to $\varnothing$ for next feature layer 
\label{algline:update-num-iterations-thus-far}
}

\vspace{0.9mm}
\Until{no new features emerge $\mathbf{or}$ the max number of layers (\emph{depth}) is reached}
\label{algline:stopping-criterion}

\vspace{1.1mm}
\State {\bf return} $\mX$ and the set of relational functions (definitions) $\F$
\label{algline:framework-return-X-and-composed-graph-functions-F}
\end{algorithmic}
\end{spacing}
\vspace{1.12mm}
}
\end{algorithm}
\end{minipage}
\end{figure*}

\subsection{Searching the Space of Relational Functions}\label{sec:searching-the-space-of-rel-functions}
A general and flexible framework for $\deepGL$ is given in Alg.~\ref{alg:framework}.
Recall that $\deepGL$ begins with a set of base features and uses these as a basis for learning deeper and more discriminative features of increasing complexity (Line~\ref{algline:base-features}).
The base feature vectors are then transformed if needed (Line~\ref{algline:transform-base-features-logarthmic-binning}).\footnote{
For instance, one may transform each feature vector $\vx_i$ using logarithmic binning as follows: 
sort $\vx_i$ in ascending order and set the $\alpha\m$ graph elements (edges/nodes) with smallest values to $0$, then set the remaining $\alpha$ graph elements to 1, and so on.}
Many normalization schemes and other techniques exist for transforming the feature vectors appropriately.
However, transformation of the feature vectors in Line~\ref{algline:transform-base-features-logarthmic-binning} \emph{and} Line~\ref{algline:transform-feature-layer-logarthmic-binning} of Alg.~\ref{alg:framework} are optional and depends on various factors.

The framework proceeds to learn a hierarchical graph representation where each successive layer represents increasingly deeper higher-order (edge/node) graph functions (due to composition): $\F_{1} < \F_{2} < \cdots < \F_{\tau}$ $\emph{s.t.}$ if $i<j$ then $\F_{j}$ is said to be deeper than $\F_{i}$.
In particular, the feature layers $\F_2, \F_{3}, \cdots, \F_{\tau}$ are learned as follows (Alg.~\ref{alg:framework} Lines~\ref{algline:repeat-feature-layer}-\ref{algline:stopping-criterion}): 
First, we derive the feature layer $\F_{\tau}$ 
by searching over the space of graph functions that arise from applying the relational feature operators
$\mPhi$ to each of the novel features $f_i \in \F_{\tau-1}$ learned in the previous layer (Alg.~\ref{alg:framework} Line~\ref{algline:construct-candidates-and-search-features}).
An example approach is given in Alg.~\ref{alg:feature-layer}.\footnote{Note that Alg.~\ref{alg:feature-layer} can be further generalized by replacing 
$\big\lbrace\N^{+}_{\ell}\!(g_i),\,\, \N^{-}_{\ell}\!(g_i),\,\, \N_{\ell}(g_i)\big\rbrace$ in Line~\ref{algline:feature-layer-foreach-set-S-for-graph-element} by a set $\mathcal{S}$.}
Further, an intuitive example is provided in Fig.~\ref{table:rel-deep-learning-operators} (Right).
Next, the feature vectors from layer $\F_{\tau}$ are transformed in Line~\ref{algline:transform-feature-layer-logarthmic-binning} (if needed) as discussed previously.

\begin{figure}[h!]
\begin{algorithm}[H]
\caption{\;\small
Derive a feature layer using the features from the previous layer and the set of relational feature operators $\ops=\{\phi_1,\cdots,\phi_K\}$.
}
\label{alg:feature-layer}
\footnotesize
\begin{spacing}{1.4}
\fontsize{8}{9}\selectfont
\algrenewcommand{\alglinenumber}[1]{\fontsize{6.5}{7}\selectfont#1}
\begin{algorithmic}[1]
\vspace{0.5mm}
\Procedure{FeatureLayer}{$G$, $\mX$, $\ops$, $\F$, $\F_{\tau-1}$}
\parfor[{\bf each} graph element $g_i \in \mathcal{G}$] \label{algline:feature-layer-parfor-each-graph-element}
	\State Reset $t$ to $\f$ for the new graph element $g_i$ (edge, node) \label{algline:feature-layer-reset-t-for-new-graph-element}
	\For{{\bf each} feature $\vx_k$ \emph{s.t.} $f_k \in \F_{\tau-1}$ in order} \label{algline:feature-layer-for-each-feature-in-prev-layer} 
		\For{{\bf each} $S \in \big\lbrace\N^{+}_{\ell}\!(g_i),\,\, \N^{-}_{\ell}\!(g_i),\,\, \N_{\ell}(g_i)\big\rbrace$} \label{algline:feature-layer-foreach-set-S-for-graph-element}		
				\For{{\bf each} relational operator $\phi \in \ops$} \Comment{See Fig.~\ref{table:rel-deep-learning-operators}} \label{algline:feature-layer-foreach-rel-op}
						\State $X_{it} = \phi\innerS{S, \vx_k}$
						and $t \leftarrow t+1$	  \label{algline:feature-layer-compute-feature-value-via-rel-op}
				\EndFor
		\EndFor
	\EndFor	
\EndPar
\State Add feature definitions to $\F_{\tau}$ \label{algline:feature-layer-add-feature-definitions-to-Ftau}
\State \textbf{return} feature matrix $\mX$ and $\F_{\tau}$ \label{algline:feature-layer-return-X-and-Ftau}
\EndProcedure
\vspace{0.22mm}
\end{algorithmic}
\end{spacing}
\end{algorithm}
\vspace{-8.6mm}
\begin{algorithm}[H]
\caption{\;\fontsize{9}{10}\selectfont
Score and prune the feature layer
}
\label{alg:score-and-prune-feature-layer}
\footnotesize
\begin{spacing}{1.3}
\fontsize{8}{9}\selectfont
\algrenewcommand{\alglinenumber}[1]{\fontsize{6.5}{7}\selectfont#1}
\begin{algorithmic}[1]
\vspace{0.5mm}
\Procedure{EvaluateFeatureLayer}{$G$, $\mX$, $\F$, $\F_{\tau}$}

\multilinenospace{
Let $\mathcal{G}_F = (V_F, E_F, \mW)$ be the initial feature graph for feature layer $\F_{\tau}$ where $V_F$ is the set of features from $\F \cup \F_{\tau}$ and $E_F = \varnothing$ 
\label{algline:set-initial-feature-graph}
}

\vspace{1.5mm}
\ParFor[{\bf each} feature $f_i \in \F_{\tau}$] \label{algline:for-each-feature-i-in-curr-layer}
\For{{\bf each} feature $f_j \in (\F_{\tau-1} \cup \cdots \cup \F_{1})$} \label{algline:for-each-feature-j-in-prev-layers}
	\If{$\corr\big(\vx_i, \vx_j\big) > \lambda$} \label{algline:feature-scoring}
		 \State Add edge $(i, j)$ to $E_F$ with weight $W_{ij} = \corr\big(\vx_i, \vx_j\big)$ \label{algline:add-edge-between-pair-of-features}
	\EndIf
\EndFor
\EndPar
\multilinenospace{
Partition $\mathcal{G}_F$ using connected components $\mathcal{C} = \{\mathcal{C}_1, \mathcal{C}_2, \ldots\}$ \label{algline:partition-feature-graph-for-pruning}
}
\ParFor[{\bf each}  $\mathcal{C}_k \in \mathcal{C}$] \Comment{Remove features} \label{algline:for-each-conn-component-prune}
	\State Find the earliest feature $f_i$ s.t. $\forall f_j \in \mathcal{C}_k : i < j$. \label{algline:find-earliest-feature-to-keep} 
	\State Remove $\mathcal{C}_k$ from $\F_{\tau}$ and set $\F_{\tau} \leftarrow \F_{\tau} \cup \{f_i\}$ \label{algline:remove-C-and-add-novel-feature}
\EndPar
\EndProcedure
\vspace{0.2mm}
\end{algorithmic}
\end{spacing}
\end{algorithm}
\end{figure}

The resulting features in layer $\tau$ are then evaluated.
The feature evaluation routine (in Alg.~\ref{alg:framework} Line~\ref{algline:main-prune-feature-layer}) chooses the important features (relational functions) at each layer $\tau$ from the space of novel relational functions (at depth $\tau$) constructed by applying the relational feature operators to each feature (relational function) learned in the previous layer $\tau-1$.
Notice that $\deepGL$ is extremely flexible as the feature evaluation routine called in Line~\ref{algline:main-prune-feature-layer} of Alg.~\ref{alg:framework} is completely interchangeable and can be fine-tuned for specific applications and/or data.
Nevertheless, an example is provided in Alg.~\ref{alg:score-and-prune-feature-layer}. 
This approach derives a score between pairs of features.
Pairs of features $\vx_i$ and $\vx_j$ that are \emph{strongly dependent} as determined by the hyperparameter $\lambda$ and evaluation criterion $\mathbb{K}$ are assigned $W_{ij}=\mathbb{K}(\vx_i,\, \vx_j)$ and $W_{ij}=0$ otherwise\footnote{This process can be viewed as a sparsification of the feature graph.} (Alg.~\ref{alg:score-and-prune-feature-layer} Line~\ref{algline:set-initial-feature-graph}-\ref{algline:add-edge-between-pair-of-features}).
More formally, let $\E_F$ denote the set of edges representing dependencies between features:
\begin{equation} \label{eq:feature-graph-objective}
E_{F} = \big\lbrace (i, j) \; | \; \forall (i, j) \in |\F| \times |\F| \text{ \it s.t. } \corr(\vx_i, \vx_j) > \lambda \big\rbrace
\end{equation}
\noindent
The result is a \emph{weighted feature dependence graph} $\mathcal{G}_F=(V_F, E_F)$ where a relatively large edge weight $\corr(\vx_i, \vx_j)=W_{ij}$  between $\vx_i$ and $\vx_j$ indicates a potential dependence (or similarity/correlation) between these two features.
Intuitively, $\vx_i$ and $\vx_j$ are strongly dependent if $\corr(\vx_i, \vx_j)=W_{ij}$ is larger than $\lambda$.
Therefore, an edge is added between features $\vx_i$ and $\vx_j$ if they are strongly dependent.
An edge between features represents (potential) redundancy.
Now, $\mathcal{G}_F$ is used select a subset of important features from layer $\tau$.
Features are selected as follows:
First, the feature graph $\mathcal{G}_F$ is partitioned into groups of features $\{\mathcal{C}_1, \mathcal{C}_2, \ldots \}$ where each set $\mathcal{C}_k \in \mathcal{C}$ represents features that are dependent (though not necessarily pairwise dependent).
To partition the feature graph $\mathcal{G}_F$, Alg.~\ref{alg:score-and-prune-feature-layer} 
uses connected components, though other methods are also possible, \eg, a clustering or community detection method.
Next, one or more representative features are selected from each group (cluster) of dependent features.
Alternatively, it is also possible to derive a new feature from the group of dependent features, \eg, finding a low-dimensional embedding of these features or taking the principal eigenvector.
In the example given in Alg.~\ref{alg:score-and-prune-feature-layer}: 
the earliest feature in each connected component $\mathcal{C}_{k} = \{...,f_i,...,f_j,...\} \in \mathcal{C}$ is selected and all others are removed.
Recall the feature evaluation routine described above is completely interchangeable by simply replacing Line~\ref{algline:main-prune-feature-layer} (Alg.~\ref{alg:framework}) of the $\deepGL$ framework.

After pruning the feature layer $\F_{\tau}$, the discarded features are removed from $\mX$ and $\deepGL$ updates the set of features learned thus far by setting $\F\leftarrow \F \,\cup\, \F_{\tau}$ (Alg.~\ref{alg:framework}: Line~\ref{algline:removed-pruned-feats-from-X}). 
Next, Line~\ref{algline:update-num-iterations-thus-far} increments $\tau$ and sets $\F_{\tau} \leftarrow \varnothing$. 
Finally, we check for convergence, and if the stopping criterion is not satisfied, then $\deepGL$ tries to learn an additional feature layer
(Line~\ref{algline:repeat-feature-layer}-\ref{algline:stopping-criterion}).
In contrast to node embedding methods that output only a \emph{node} feature matrix $\mX$, $\deepGL$ also outputs the (hierarchical) relational functions (definitions) $\F=\{\F_1,\, \F_2,\, \cdots\,\}$ where each $\f_i \in \F_h$ is a learned relational function of depth $d$ for the $i$-th feature vector $\vx_i$.
Maintaining the relational functions are important
for transferring the features to another arbitrary graph of interest,
but also for interpreting them.

\begin{figure*}[t!]
\centering
\includegraphics[width=0.65\linewidth]{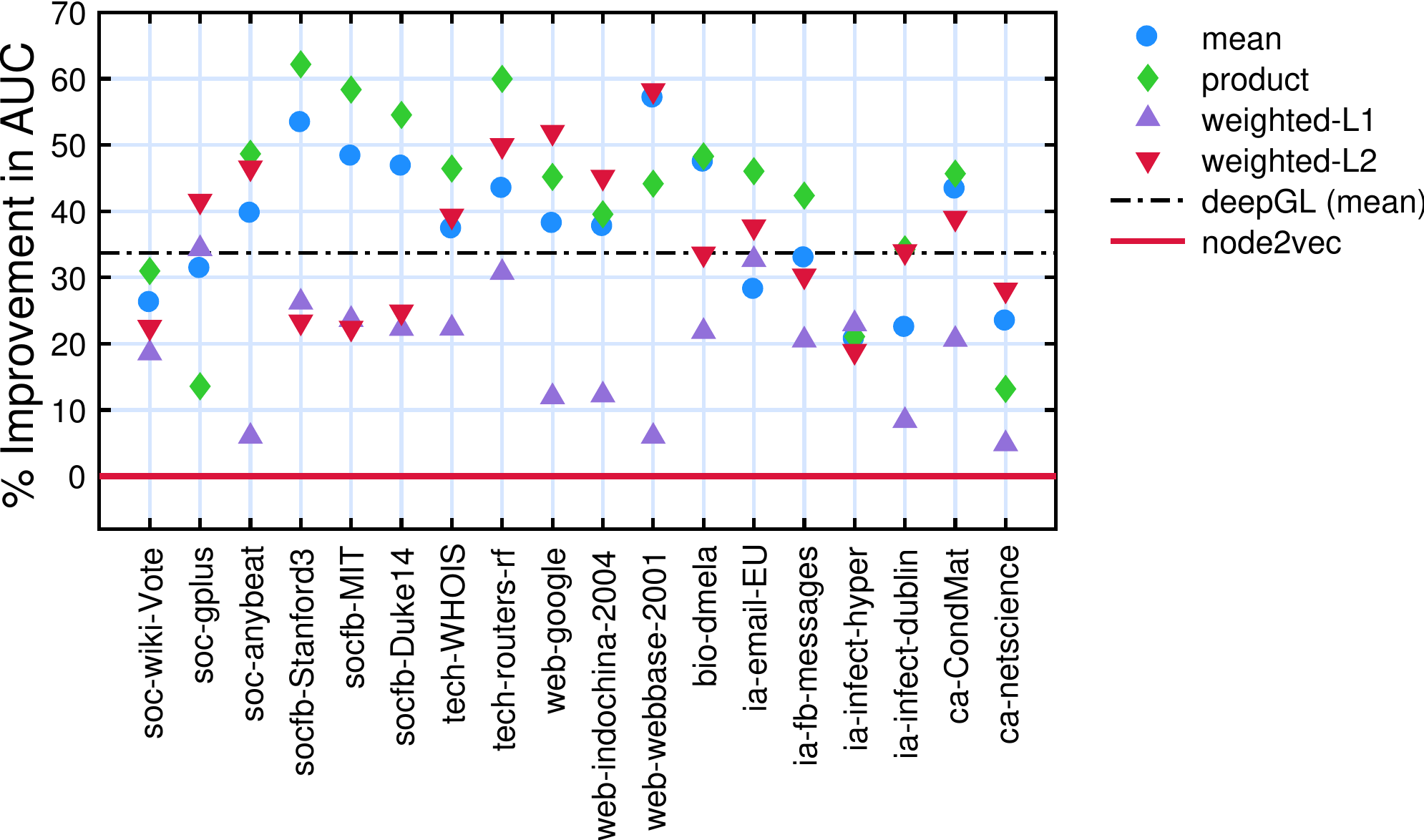}
\caption{$\deepGLsf$ is effective for link prediction with significant improvement in predictive performance over node2vec.}
\label{fig:link-pred-perc-improvement-auc-score}
\end{figure*}

\subsection{Feature Diffusion} \label{sec:framework-feature-diffusion}
We introduce the notion of feature diffusion where the feature matrix at each layer can be smoothed using any arbitrary feature diffusion process.
As an example, suppose $\mX$ is the resulting feature matrix from layer $\tau$, 
then we can set $\bar{\mX}^{(0)}\leftarrow \mX$ and solve $\bar{\mX}^{(t)} = \mD^{-1}\mA\bar{\mX}^{(t-1)}$ where $\mD$ is the diagonal degree matrix and $\mA$ is the adjacency matrix of $G$. 
The diffusion process above is repeated for a fixed number of iterations $t=1,2,...,T$ or until convergence; and $\bar{\mX}^{(t)} = \mD^{-1}\mA\bar{\mX}^{(t-1)}$ corresponds to a simple feature propagation.
More complex feature diffusion processes can also be used in $\deepGL$ such as the normalized Laplacian feature diffusion defined as 
\begin{equation}
\bar{\mX}^{(t)} = (1-\theta)\mL\bar{\mX}^{(t-1)} + \theta\mX,\quad \text{  for } t=1,2,...
\end{equation}
\noindent
where $\mL$ is the normalized Laplacian:
\begin{equation}
\mL = \eye - \mD^{\nicefrac{1}{2}}\mA\mD^{\nicefrac{1}{2}}
\end{equation} 
The resulting diffused feature vectors $\bar{\mX} = \big[\;\; \bar{\vx}_1 \;\; \bar{\vx}_2 \;\; \cdots \;\; \big]$ are effectively smoothed by the features of related graph elements (nodes/edges) governed by the particular diffusion process.
Notice that feature vectors given as output at each layer can be diffused (\eg, after Line~\ref{algline:construct-candidates-and-search-features} or~\ref{algline:removed-pruned-feats-from-X} of Alg.~\ref{alg:framework}).
The resulting features $\bar{\mX}$ can be leveraged in a variety of ways.
For instance, one can set $\mX \leftarrow \bar{\mX}$ and thereby replacing the existing features with the diffused versions, or alternatively, the diffused features can be added to $\mX$ by setting 
$\mX \leftarrow \big[\, \mX \; \bar{\mX} \,\big]$.
Further, the diffusion process can be learned via cross-validation.

\subsection{Supervised Graph Representation Learning}
\label{sec:framework-supervised-graph-rep-learning}
The $\deepGL$ framework naturally generalizes for \emph{supervised representation learning} by replacing the feature evaluation routine (called in Alg.~\ref{alg:framework} Line~\ref{algline:main-prune-feature-layer}) with an appropriate objective function, \eg, one that seeks to find a set of features that 
($i$) maximize relevancy (predictive quality) with respect to $\vy$ (\ie, observed class labels) while 
($ii$) minimizing redundancy between each feature in that set.
The objective function capturing both ($i$) and ($ii$) can be formulated by replacing $\mathbb{K}$ with a measure such as mutual information (and variants):
\begin{equation} \label{eq:supervised-graph-rep-learning}
\vx = \argmax_{\vx_i \not\in \mathcal{X}} \; \Bigg\lbrace \mathbb{K} \big(\vy,\, \vx_i\big) - \beta \sum_{\vx_j \in \mathcal{X}} \mathbb{K} \big( \vx_i,\, \vx_j\big) \Bigg\rbrace
\end{equation}
\noindent 
where $\mathcal{X}$ is the current set of selected features; and 
$\beta$ is a hyperparameter that determines the balance between maximizing relevance \emph{and} minimizing redundancy.
The first term in Eq.~\eqref{eq:supervised-graph-rep-learning} seeks to find $\vx_i$ that maximizes the relevancy of $\vx_i$ to $\vy$ whereas the second term attempts to minimize the redundancy between $\vx_i$ and each $\vx_j \in \mathcal{X}$ of the already selected features.
Initially, 
\begin{equation}
\mathcal{X} \leftarrow \{ \vx^{\prime} \}
\end{equation}
\noindent
where
\begin{equation}
\vx^{\prime} = \argmax_{\vx_i}
\,\, \mathbb{K} \big(\vy,\, \vx_i\big)
\end{equation}
\noindent
Afterwards, wesolve Eq.~\eqref{eq:supervised-graph-rep-learning} to find $\vx_i$ (such that $\vx_i \not\in \mathcal{X}$) which is then added to $\mathcal{X}$ (and removed from the set of remaining features).
This is repeated until the stopping criterion is reached (\eg, until the desired $|\mathcal{X}|$).
$\deepGL$ naturally supports many other objective functions and optimization schemes.

\subsection{Computational Complexity} \label{sec:framework-complexity}
Recall that $\m$ is the number of edges, $\n$ is the number of nodes, and $\nF$ is the number of features.
The total computational complexity of the \emph{edge representation learning} from the $\deepGL$ framework is 
\begin{equation} \label{eq:time-complexity-deepGL-edge}
\mathcal{O}\big(\nF(\m+\m\nF)\big)
\end{equation}
\noindent
For learning \emph{node representations} with the $\deepGL$ framework it takes $\mathcal{O}(\nF(\m+\n\nF))$.
Thus, in both cases, the runtime of $\deepGL$ is linear in the number of edges.
As an aside, the initial graphlet features are computed using fast and accurate estimation methods, see~\cite{ahmed16bigdata,rossi17graphlet-est}.

\section{Experiments} \label{sec:exp}
This section demonstrates the effectiveness of the proposed framework.

\subsection{Experimental settings} \label{sec:exp-setup}
In these experiments, we use the following instantiation of $\deepGL$:
Features are transformed using logarithmic binning and 
evaluated using a simple agreement score function where $\corr(\vx_i, \vx_j)=$ fraction of graph elements 
that agree.
The specific model from the space of models defined by the above instantiation of $\deepGL$ is selected using 10-fold cross-validation on $10\%$ of the labeled data.
Experiments are repeated for 10 random seed initializations.
All results are statistically significant with p-value $< 0.01$.

Despite the fundamental differences (in terms of problem and potential applications, see summary of differences in Section~\ref{sec:intro}) between $\deepGL$ and the recent node embedding methods such as node2vec, we evaluate the proposed framework against node2vec\footnote{\href{https://github.com/aditya-grover/node2vec}{$\mathtt{https}$://$\mathtt{github.com/aditya}$-$\mathtt{grover/node2vec}$}} whenever applicable.
For node2vec, we use the hyperparameters and grid search over $p,q\in \{0.25, 0.50, 1, 2, 4\}$ as mentioned in~\cite{node2vec}.
Results for DeepWalk~\cite{deepwalk}, LINE~\cite{line}, and spectral clustering were removed for brevity since node2vec was shown in~\cite{node2vec} to outperform these methods.
Unless otherwise mentioned, we use logistic regression with an L2 penalty and one-vs-rest strategy for multiclass problems.
For evaluation, we use AUC and Total-AUC~\cite{totalAUC-hand2001simple} for multiclass problems.
Data has been made available at NetworkRepository~\cite{nr}.\footnote{See \href{http://networkrepository.com/}{$\mathtt{http}$://$\mathtt{networkrepository.com/}$} for data details and stats.}

\subsection{Effectiveness on Link Prediction} 
\label{sec:exp-link-prediction}
Given a graph $G$ with a fraction of missing edges, the link prediction task is to predict these missing edges.
We generate a labeled dataset of edges as done in~\cite{node2vec}.
Positive examples are obtained by removing $50\%$ of edges randomly, whereas \emph{negative examples} are generated by randomly sampling an equal number of node pairs that are not connected with an edge, \ie, each node pair $(i,j) \not\in E$. 
For each method, we learn features using the remaining graph that consists of only positive examples.
Using the feature representations from each method, we then learn a model to predict whether a given edge in the test set exists in $E$ or not.
Notice that node embedding methods such as node2vec require that each node in $G$ appear in at least one edge in the training graph (\ie, the graph remains connected), otherwise these methods are unable to derive features for such nodes.\footnote{A significant limitation prohibiting the use of these methods for many applications.}

The gain/loss in predictive performance over node2vec is summarized in Fig.~\ref{fig:link-pred-perc-improvement-auc-score}.
In all cases, $\deepGL$ achieves better predictive performance over node2vec across a wide variety of graphs with different characteristics and binary operators.
For comparison, we use the same set of binary operators 
to construct features for the edges \emph{indirectly} using the learned node representations: 
$(\vx_i + \vx_j)\big/2$ is the \textsc{mean}; $\vx_i \odot \vx_j$ is the (Hadamard) \textsc{product}; $\abs{\vx_i - \vx_j}$ and $(\vx_i - \vx_j)^{\circ 2}$ is the \textsc{weighted}-\textsc{l}$_{1}$ and \textsc{weighted}-\textsc{l}$_{2}$ binary operators, respectively.\footnote{Note $\vx^{\circ 2}$ is the element-wise Hadamard power; $\vx_i \odot\, \vx_j$ is the element-wise product.}
Strikingly, $\deepGL$ improves over node2vec by up to $60\%$ and always by at least $5\%$ with an average improvement of $33.6\%$ across all graphs and binary operators.
Overall, the product and mean binary operators give the best results with an average gain in AUC of $41.9\%$ and $37.6\%$ (over all graphs), respectively.

\begin{table}[h!]
\centering
\caption{AUC scores for \emph{within-network link classification}.
The method that performs best for each graph is bold.
We also highlight the method with largest AUC score for each binary op (\eg, $\vx_i \odot \vx_j$ is the Hadamard product).
See text for discussion.
}
\label{table:link-classification-results-auc}
\setlength{\tabcolsep}{4pt} 
\centering 
\small
\def\arraystretch{1.29}
\scalebox{0.84}{
\begin{tabularx}
{2.80in}
{@{} 
c
@{}
X
X
X
HHH@{}
@{}}
\toprule
\def\arraystretch{1.0}

& & \dataName{escorts}   & 
\dataName{yahoo\text{-}msg}  & 
\\ 
\midrule

\multirow{2}{*}{\rotatebox{0}{\scriptsize $\big(\vx_i + \vx_j\big)\big/2\quad$}}
&  
\textbf{\deepGLbold}  &
\cellcolor{\verylightgreen}\text{$\mathbf{0.6891}$} &
\cellcolor{\verylightgreen}\text{$\mathbf{0.9410}$} &
\\

&  
\textbf{node2vec}  &
0.6426 &
0.9397 &
\\
\midrule

\multirow{2}{*}{\rotatebox{0}{\scriptsize $\vx_i \odot \vx_j$}}
&  
\textbf{\deepGLbold}  &
\cellcolor{\verylightgreen}\text{0.6339} &
\cellcolor{\verylightgreen}\text{0.9324} &
\\

&  
\textbf{node2vec}  &
0.5445 &
0.8633 &
\\
\midrule

\multirow{2}{*}{\rotatebox{0}{\scriptsize $\abs{\vx_i - \vx_j}$}}
&  
\textbf{\deepGLbold}  &
\cellcolor{\verylightgreen}\text{0.6857} &
\cellcolor{\verylightgreen}\text{0.9247} &
\\

&  
\textbf{node2vec}  &
0.5050 &
0.7644 &
\\
\midrule

\multirow{2}{*}{\rotatebox{0}{\scriptsize $(\vx_i - \vx_j)^{\circ 2}$}}
&  
\textbf{\deepGLbold}  &
\cellcolor{\verylightgreen}\text{0.6817} &
\cellcolor{\verylightgreen}\text{0.9160} &
\\

&  
\textbf{node2vec}  &
0.4950 &
0.7623 &
\\

\bottomrule
\end{tabularx}
}
\end{table}

\subsection{Within-Network Link Classification}
\label{sec:exp-link-classification}
Besides predicting the existence of links, we also evaluate $\deepGL$ for link classification.
To be able to compare to node2vec and other methods, we focus in this section on within-network link classification.\footnote{Recall that node2vec and other existing node embedding approaches require the training graph to contain at least one edge among each node in $G$.}
In Table~\ref{table:link-classification-results-auc}, we observe that $\deepGL$ outperforms node2vec in all graphs with a gain in AUC of up to $7.2\%$ when using the best operator for each method.
Other results were omitted due to space.

\begin{figure}[h!]
\centering
\scalebox{1.0}{
\includegraphics[width=0.75\linewidth]{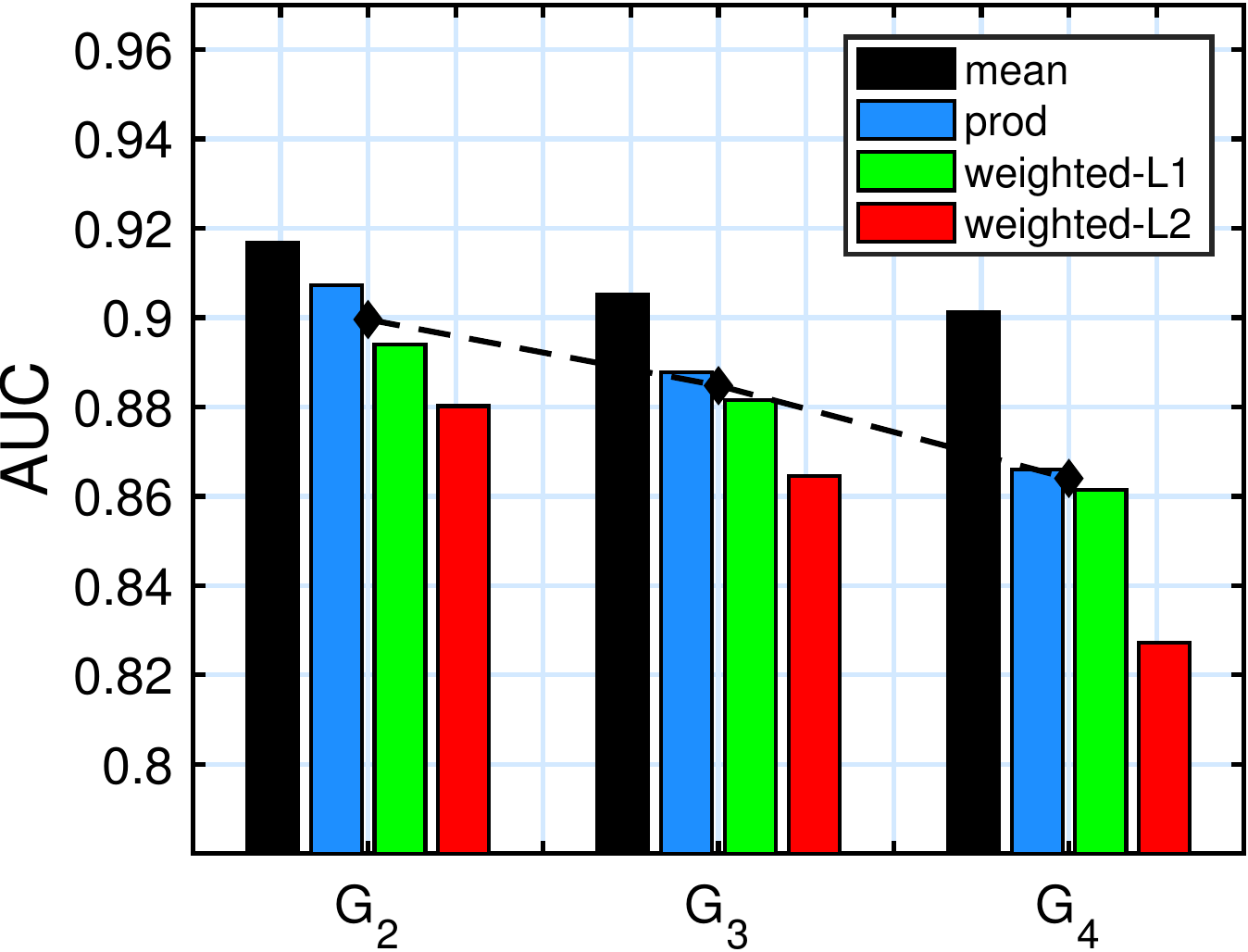}
}
\caption{
Effectiveness of $\deepGLsf$ framework for across network transfer learning.
AUC scores for across network link classification using \dataName{yahoo\text{-}msg}.
Note ${\blacklozenge}$ denotes the mean AUC of each test graph.
}
\label{fig:exp-across-network-classification}
\end{figure}

\subsection{Graph-based Transfer Learning}
\label{sec:exp-transfer-learning}
Recall from Section~\ref{sec:framework} that a key advantage of $\deepGL$ (over existing methods such as~\cite{node2vec,deepwalk,line}) lies in its ability to learn features that naturally generalize for across-network transfer learning tasks.
In particular, the features learned by $\deepGL$ are fundamentally different than existing methods 
as they represent a composition (or convolution) of one or more base relational feature operators applied to an initial set of base graph features that are easily computed on any arbitrary graph.

For each experiment, the training graph is fully observed with all known labels available for learning.
The test graph is completely unlabeled and each classification model is evaluated on its ability to predict \emph{all} available labels in the test graph.
Given the training graph $G=(V,E)$, we use $\deepGL$ to learn the feature matrix $\mX$
and the relational functions $\F$ (definitions).
The relational functions $\F$ are then used to extract the same identical features on an arbitrary \emph{test graph} $G^{\prime}=(V^{\prime}, E^{\prime})$ giving as output a feature matrix $\mX^{\prime}$.\footnote{Notice that each node (or edge) is embedded in the same $\nF$-dimensional space, even despite that the set of nodes/edges between the graphs could be completely disjoint.}
Thus, an identical set of features is used for all train and test graphs.

\begin{figure}[h!]
\centering
\includegraphics[width=0.75\linewidth]{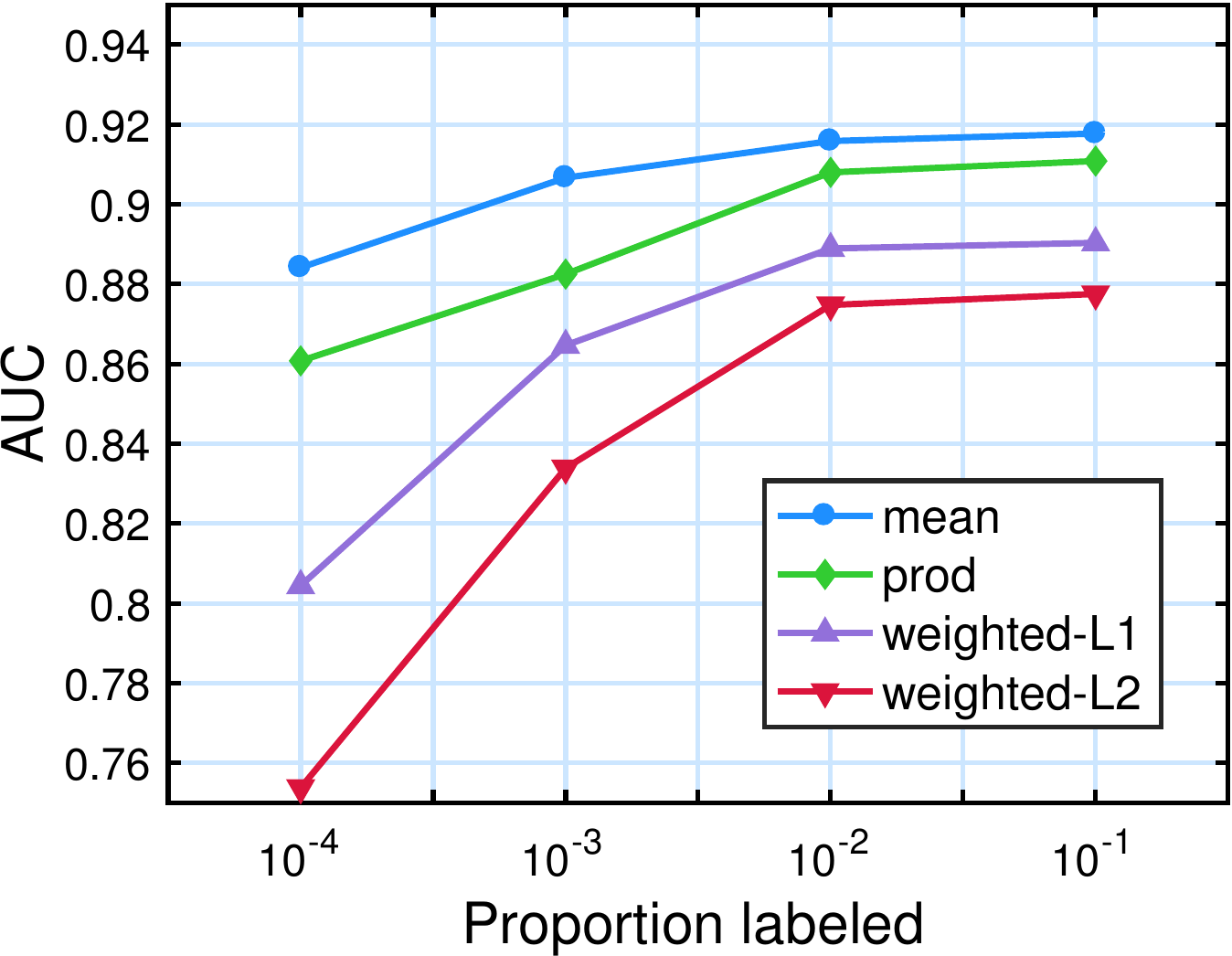} 
\caption{
Effectiveness of $\deepGLsf$ for link classification with very small amounts of training labels. 
}
\label{fig:exp-transfer-learning-vary-training}
\end{figure}

In these experiments, the training graph $G_1$ represents the first week of data from $\dataName{yahoo\text{-}msg}$,\footnote{\href{https://webscope.sandbox.yahoo.com/}{$\mathtt{https}$://$\mathtt{webscope.sandbox.yahoo.com/}$}} whereas the test graphs 
$\{G_2,G_3,G_4\}$ represent the next three weeks of data (\eg, $G_2$ contains edges that occur only within week 2, 
and so on).
Hence, the test graphs contain many nodes and edges not present in the training graph.
As such, the predictive performance is expected to decrease significantly over time as the features become increasingly stale due to the constant changes in the graph structure with the addition and deletion of nodes and edges.
However, we observe the performance of $\deepGL$ for across-network link classification to be stable with only a small decrease in AUC as a function of time as shown in Fig.~\ref{fig:exp-across-network-classification}. 
This is especially true for edge features constructed using mean.
As an aside, the mean operator gives best performance on average across all test graphs; with an average AUC of 0.907 over all graphs.

Now we investigate the performance as a function of the amount of labeled data used.
In Fig.~\ref{fig:exp-transfer-learning-vary-training}, we observe that $\deepGL$ performs
well with very small amounts of labeled data for training.
Strikingly, the difference in AUC scores from models learned using $1\%$ of the labeled data is insignificant at $p<0.01$ $\wrt$ models learned using larger quantities.

\begin{table}[h!]
\centering
\setlength{\tabcolsep}{5.5pt}
\ra{1.1}
\caption{Node classification results for binary and multiclass problems.}
\vspace{1mm}
\label{table:eval-methods-auc-node-classification}
\footnotesize
\begin{tabularx}{0.7\linewidth}
{@{}r Hc cc 
H
@{}}
\hline
\TTT\BB && 
\multicolumn{3}{c}{\normalsize { \sc auc }} &
\\
\cmidrule(l{-2pt}r{5pt}){4-6} 

\textbf{graph}  && 
$C$   &   
$\deepGLbold$
& \textbf{node2vec} 
\\
\midrule

\datasmone{DD242}  && 20  & 
\textbf{0.730} &   0.673 
\\ 

\datasmone{DD497}  && 20  & 
\textbf{0.696} &   0.660 
\\ 

\datasmone{DD68}  && 20  & 
\textbf{0.730} &    0.713 
\\ 

\datasmone{ENZYMES118}  &&  2  &  
\textbf{0.779} &   0.610 
\\ 

\datasmone{ENZYMES295}  && 2  & 
\textbf{0.872} &  0.588 
\\ 

\datasmone{ENZYMES296}  && 2  &
\textbf{0.823} &   0.610 
\\ 

\bottomrule
\end{tabularx}
\end{table}

\subsection{Node Classification} \label{sec:exp-node-classification}
For node classification, we use the $\iid$ variant of $\rsm$~\cite{rossi16rsm} since it is able to handle multiclass problems in a direct fashion (as opposed to indirectly, \eg, one-vs-rest) and consistently outperformed other indirect approaches such as LR and SVM.
In particular, $\rsm$ assigns a test vector $\vx_i$ to the class that is most similar $\wrt$
the training vectors (\ie, feature vectors of the nodes with known labels); see~\cite{rossi16rsm} for further details.
Similarity is measured using the RBF kernel and RBF's hyperparameter $\sigma$ is set using cross-validation with a grid search over $\sigma \in \{0.001,0.01,0.1,1\}$.
Results are shown in Table~\ref{table:eval-methods-auc-node-classification}.
In all cases, we observe that $\deepGL$ significantly outperforms node2vec across all graphs and node classification problems including both binary and multiclass problems.
Further, $\deepGL$ achieves the best improvement in AUC on ENZYMES295 of $48\%$.

\begin{figure}[h!]
\includegraphics[width=0.9\linewidth]{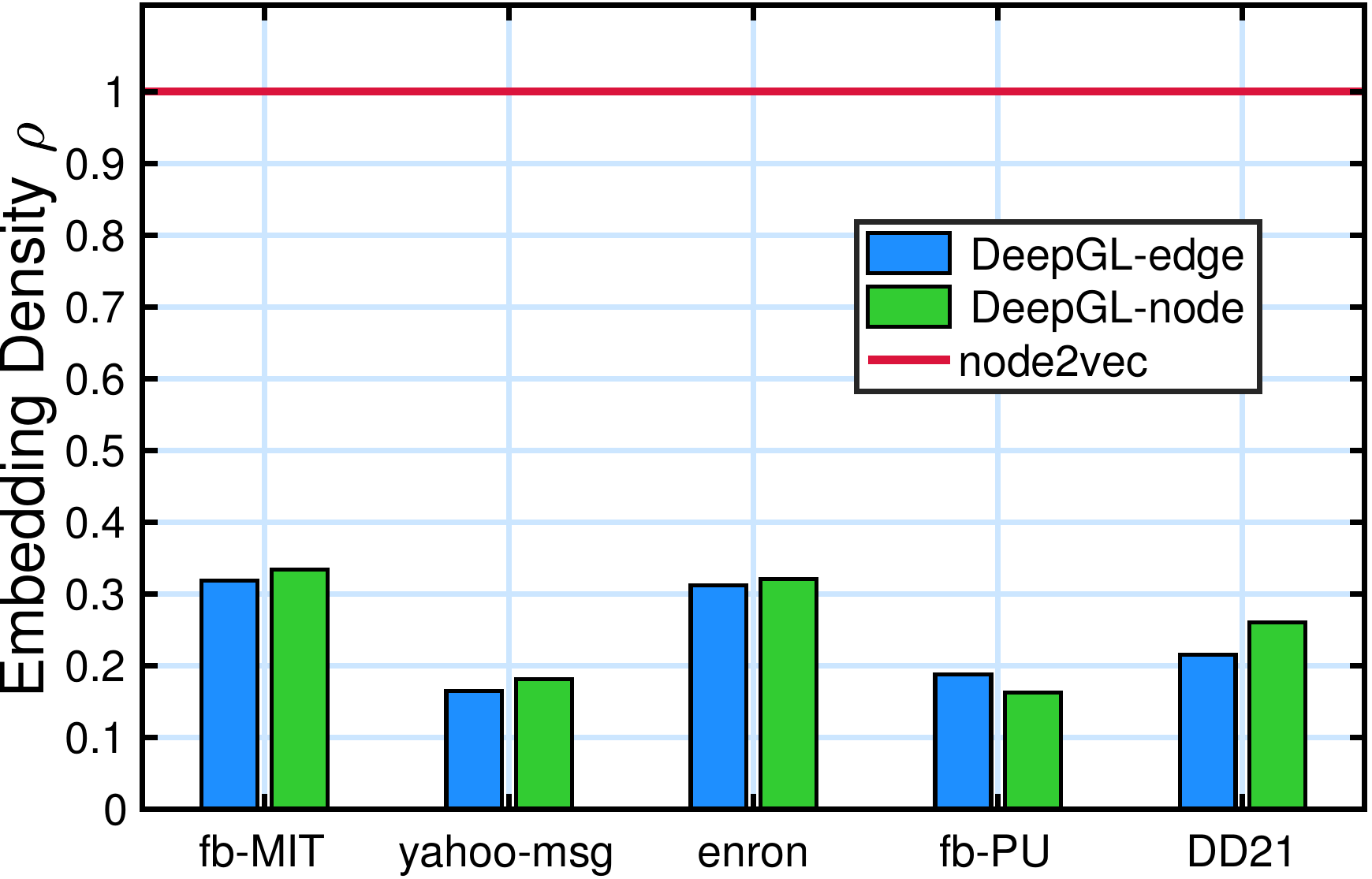} 
\caption{
Comparing the sparsity of learned features.
Notably, $\deepGLsf$ is space-efficient and uses up to $6$x less space than existing methods.
See text for discussion.
}
\label{fig:exp-sparse-graph-rep}
\end{figure}

\subsection{Analysis of Space-Efficiency} 
\label{sec:exp-sparse-graph-rep-learning}
Learning sparse space-efficient node and edge feature representations is of vital importance for large networks where storing even a modest number of \emph{dense} features is impractical (especially when stored in-memory).
Despite the importance of learning a sparse space-efficient representation, existing work has been limited to discovering completely dense (node) features~\cite{deepwalk,node2vec,line}.
To understand the effectiveness of the proposed framework for learning sparse graph representations, we measure the density of each representation learned from $\deepGL$ and compare these against the state-of-the-art methods~\cite{node2vec,deepwalk}.
We focus first on node representations since existing methods are limited to only node features.
Results are shown in Fig.~\ref{fig:exp-sparse-graph-rep}.
In all cases, the node representations learned by $\deepGL$ are extremely sparse and significantly more space-efficient than node2vec~\cite{node2vec} as observed in Fig.~\ref{fig:exp-sparse-graph-rep}.
Strikingly, $\deepGL$ uses only a fraction of the space required by existing methods 
(Fig.~\ref{fig:exp-sparse-graph-rep}).
Moreover, the density of node and edge 
representations from $\deepGL$ is between $\big[\begin{smallmatrix}0.162, & 0.334\end{smallmatrix}\big]$ for nodes and $\big[\begin{smallmatrix}  0.164, & 0.318\end{smallmatrix}\big]$ for edges and up to $6\times$ more space-efficient than existing methods.

\begin{figure}[t!]
\centering
\subfigure[Runtime comparison]
{\includegraphics[width=0.75\linewidth]
{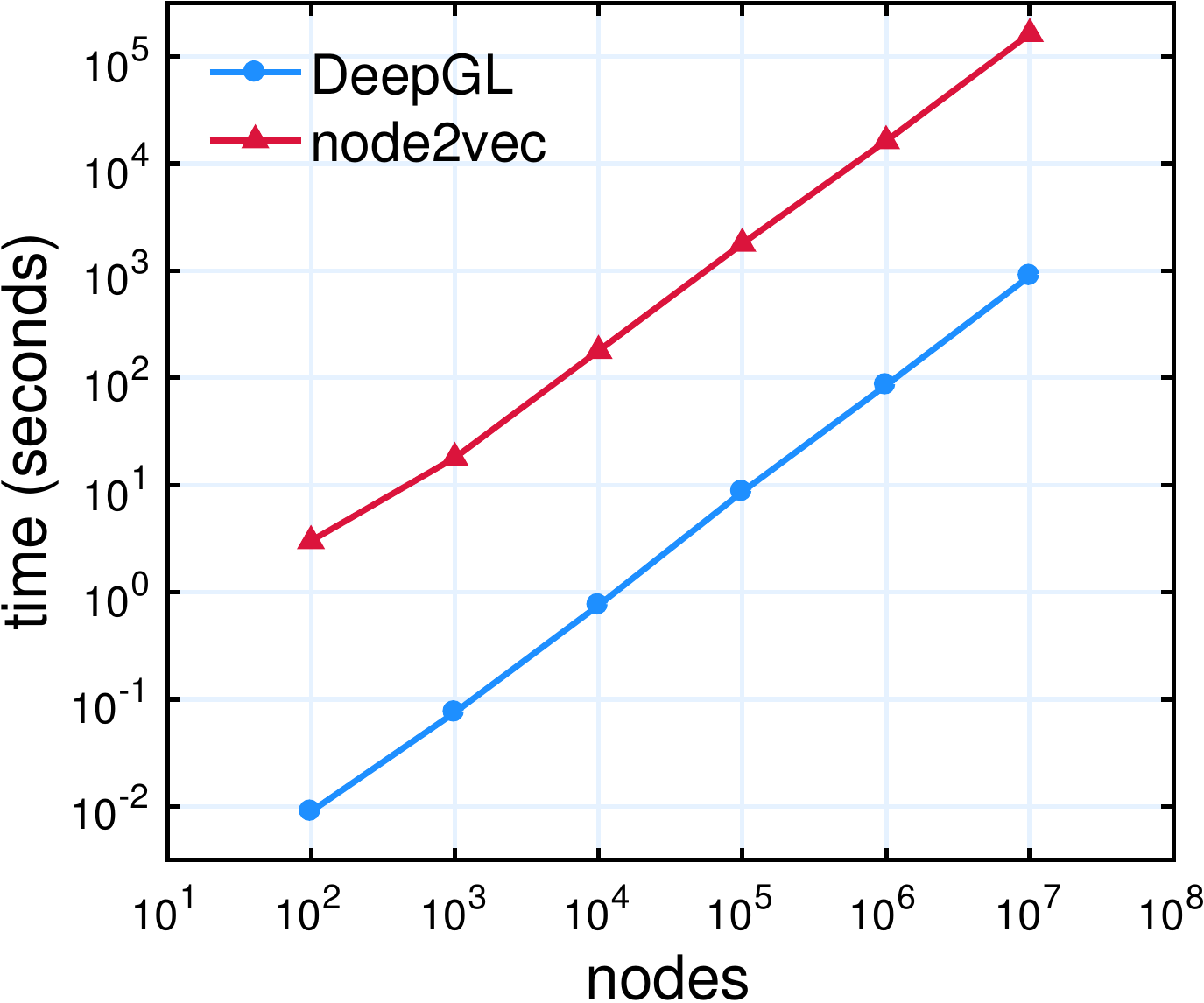}\label{fig:exp-runtime-ER-ours-vs-node2vec}}
\subfigure[Runtime of phases]
{\includegraphics[width=0.75\linewidth]
{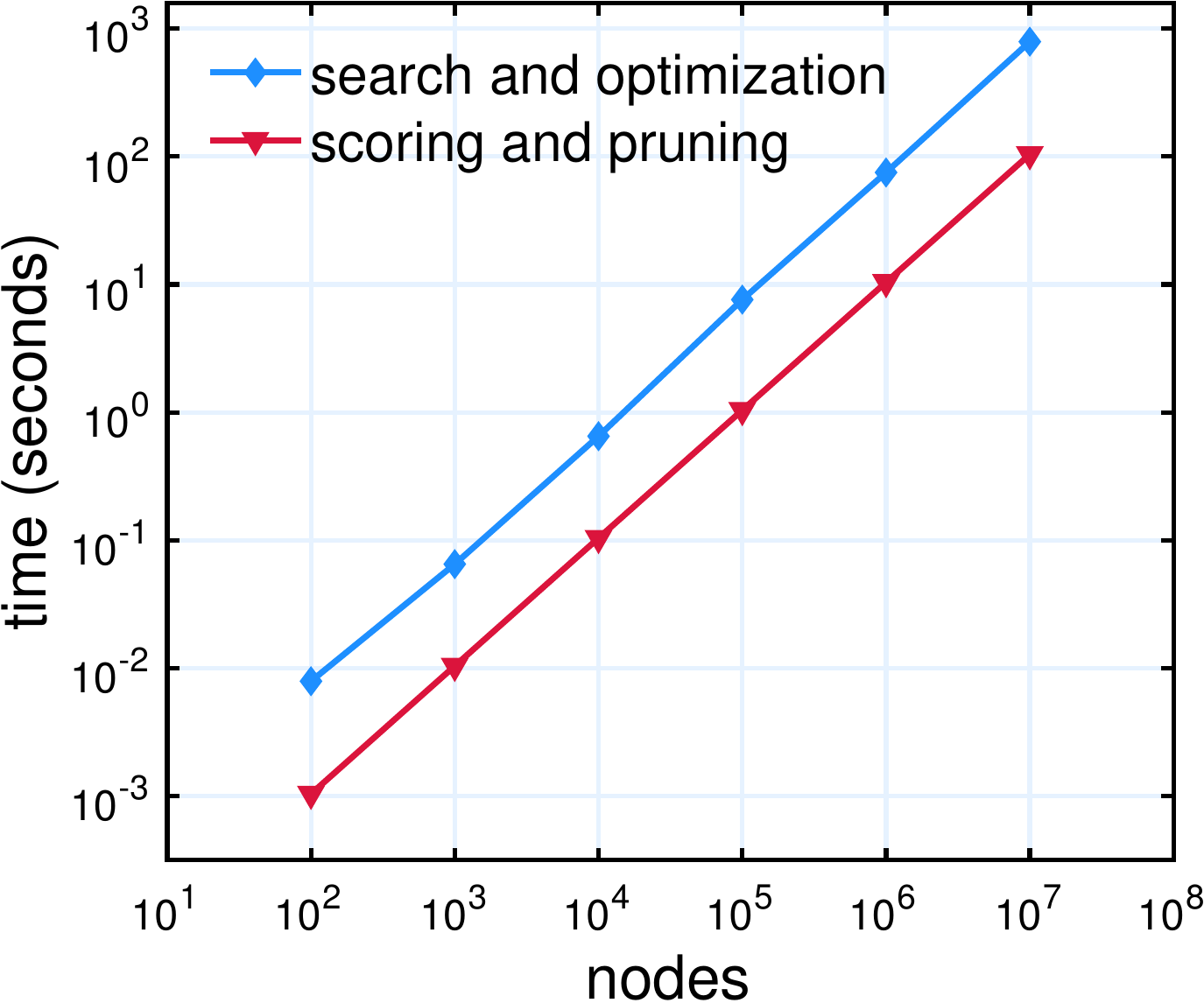}\label{fig:exp-runtime-ER-search-vs-pruning-and-scoring}}
\caption{
Runtime comparison on  Erd\"{o}s-R\'{e}nyi graphs with an average degree of 10.
\textit{(a)} 
The proposed approach is shown to be orders of magnitude faster than node2vec~\cite{node2vec}.
\textit{(b)} 
Runtime of the main $\deepGLsf$ phases.
}
\label{fig:exp-runtime-ER}
\end{figure}

Notably, recent node embedding methods not only output dense node features, 
but are also real-valued and often negative (\eg,~\cite{deepwalk,node2vec,line}).
Thus, they require 8 bytes per feature-value, whereas $\deepGL$ requires only 2 bytes and can sometimes be reduced to even 1 byte if needed by adjusting $\alpha$ (\ie, the bin size of the log binning transformation).
To understand the impact of this, assume both approaches learn a node representation with 128 dimensions (features) for a graph with 10,000,000 nodes.
In this case, node2vec requires $10.2$GB, whereas $\deepGL$ uses only $0.768$GB (assuming a modest $0.3$ density) --- a significant reduction in space by a factor of 13.

\subsection{Runtime \& Scalability} \label{sec:exp-runtime-scalability}
To evaluate the performance and scalability of the proposed framework, we learn node representations for Erd\"{o}s-R\'{e}nyi graphs of increasing size (from 100 to 10,000,000 nodes) such that each graph has an average degree of 10.
We compare the performance of $\deepGL$ against node2vec~\cite{node2vec} -- a recent node embedding method based on DeepWalk~\cite{deepwalk} that is specifically designed to be \emph{scalable}.
Default parameters are used for each method.
In Fig.~\ref{fig:exp-runtime-ER-ours-vs-node2vec}, we observe that $\deepGL$ is significantly faster and more scalable than node2vec.
In particular, node2vec takes 1.8 days (45.3 hours) for 10 million nodes, whereas $\deepGL$ finishes in only 15 minutes; see Fig.~\ref{fig:exp-runtime-ER-ours-vs-node2vec}.
Strikingly, this is $182$ times faster than node2vec.
In Fig.~\ref{fig:exp-runtime-ER-search-vs-pruning-and-scoring}, we observe that $\deepGL$ spends the majority of time in the search and optimization phase. 

\begin{figure}[h!]
\centering
\includegraphics[width=0.64\linewidth]{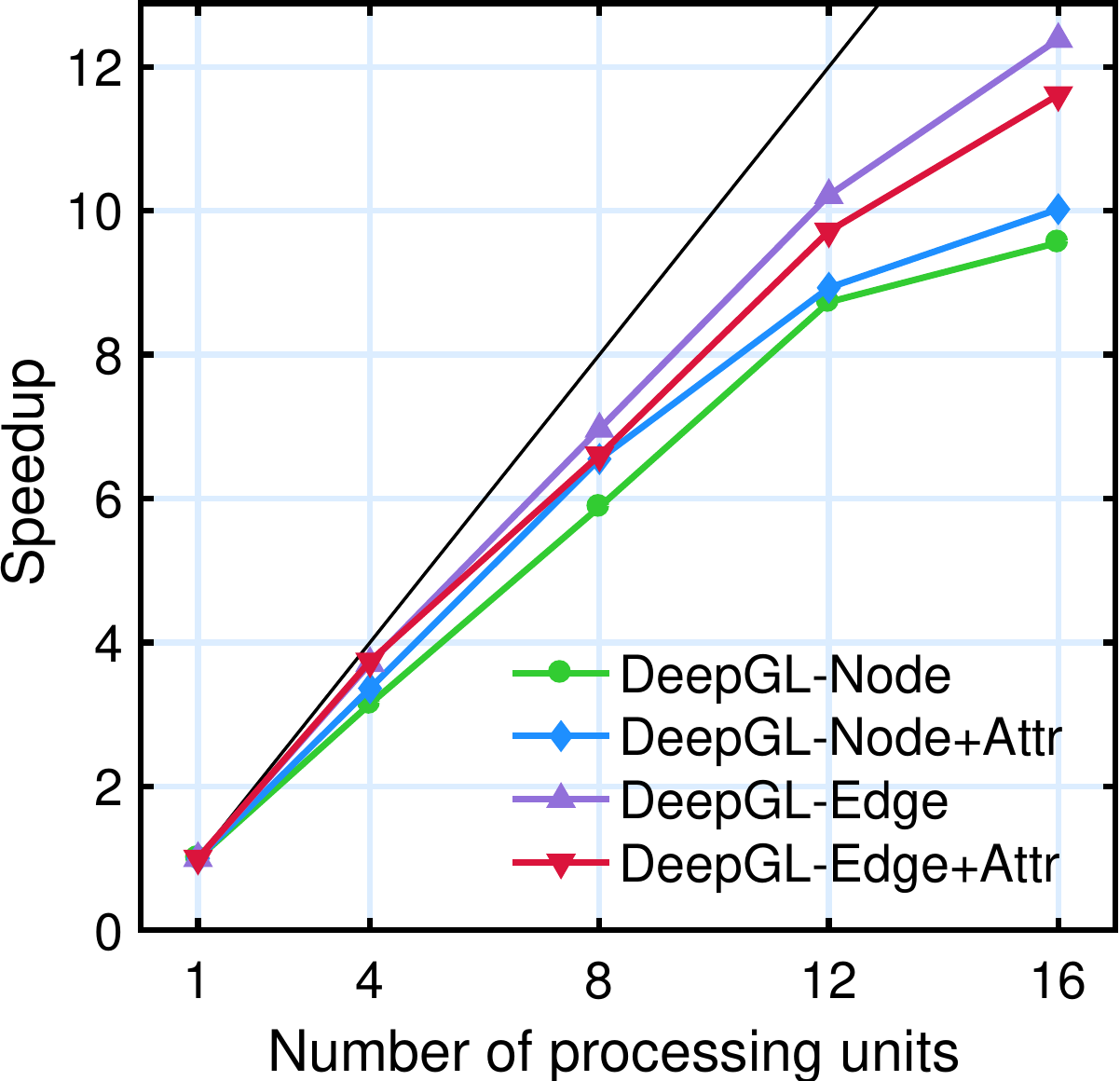}
\caption{Parallel speedup of different $\deepGLsf$ variants.
See text for discussion.
}
\label{fig:exp-parallel-scaling}
\end{figure}

\subsection{Parallel Scaling} \label{sec:exp-parallel-scaling}
This section investigates the parallel performance of $\deepGL$.
In Fig.~\ref{fig:exp-parallel-scaling}, we observe strong parallel scaling for all $\deepGL$ variants with the edge representation learning variants 
performing slightly better than the node representation learning methods from $\deepGL$.
Results are reported for $\data{soc}{gowalla}$ on a machine with 4 Intel Xeon E5-4627 v2 3.3GHz CPUs.
Similar results were found for other graphs and machines.

\section{Conclusion} \label{sec:conc}
We propose $\deepGL$, a general, flexible, and highly expressive framework for learning deep node and edge features from large (attributed) graphs.
Each feature learned by $\deepGL$ corresponds to a composition of relational feature operators applied over a base feature.
Thus, features learned by $\deepGL$ are interpretable and naturally generalize for across-network transfer learning tasks as they can be derived on any arbitrary graph.
The framework is flexible with many interchangeable components, expressive, interpretable, parallel, and is both space- and time-efficient for large graphs with runtime that is linear in the number of edges.
$\deepGL$ has all the following desired properties:
\smallskip
\begin{itemize}
\item \textbf{Effective} for attributed graphs and across-network transfer learning tasks
\item \textbf{Space-efficient} requiring up to $6\times$ less memory
\item \textbf{Fast} with up to $182\times$ speedup in runtime 
\item \textbf{Accurate} with a mean improvement of $20\%$ or more on many applications
\item \textbf{Parallel} with strong scaling results.
\end{itemize}

\balance
\bibliographystyle{IEEEtran}
\bibliography{paper}

\end{document}